\documentclass[10pt,journal,compsoc]{IEEEtran}
\usepackage{booktabs}
\usepackage{algorithm}
\usepackage{setspace}
\usepackage{amssymb}
\usepackage{multirow}
\usepackage{url}
\usepackage{soul}	

\usepackage{tabulary}
\usepackage{tabularx}
\usepackage{tabu}

\usepackage[capitalise]{cleveref}

\renewcommand{\ul}[1]{#1}

\ifCLASSOPTIONcompsoc
  \usepackage[nocompress]{cite}
\else
  \usepackage{cite}
\fi
\ifCLASSINFOpdf
  \usepackage[pdftex]{graphicx}
\else
\fi
\usepackage{amsmath}
\usepackage{algorithmic}
\begin{document}
%
\title{TapLab: A Fast Framework for Semantic Video Segmentation Tapping into Compressed-Domain Knowledge}
%
%
%
%

\author{Junyi~Feng, 
        Songyuan~Li, 
        Xi~Li, 
        Fei~Wu, 
        Qi~Tian, 
        Ming-Hsuan~Yang, 
        and~Haibin~Ling
\IEEEcompsocitemizethanks{\IEEEcompsocthanksitem J.~Feng,~S.~Li, X.~Li, and F.~Wu are with  College of Computer Science and Technology, Zhejiang University, Hangzhou 310027, China.\protect
  E-mail: \{fengjunyi,~leizungjyun,~xilizju\}@zju.edu.cn, wufei@cs.zju.edu.cn. \protect
  \IEEEcompsocthanksitem Q.~Tian is with Huawei Noah’s Ark Lab, Beijing 100032, China  \protect.
  E-mail: tian.qi1@huawei.com. \protect
\IEEEcompsocthanksitem M.~Yang is with Electrical Engineering and Computer Science, the University of California at Merced, Merced, CA 95344. \protect\\
  E-mail: mhyang@ucmerced.edu. \protect
\IEEEcompsocthanksitem H.~Ling is with the Department of Computer Science, Stony Brook University, Stony Brook, NY 11794-2424. \protect
  E-mail: haibin.ling@gmail.com.}

\thanks{(Corresponding author: Xi Li.)}} 

\IEEEtitleabstractindextext{%
\begin{abstract}
Real-time semantic video segmentation is a challenging task due to the strict requirements of inference speed.
Recent approaches mainly devote great efforts to reducing the model size for high efficiency. 
In this paper, we rethink this problem from a different viewpoint: using knowledge contained in compressed videos. We propose a simple and effective framework, dubbed TapLab, to tap into resources from the compressed domain.
Specifically, we design a fast feature warping module using motion vectors for acceleration. To reduce the noise introduced by motion vectors, we design a residual-guided correction module and a residual-guided frame selection module using residuals.
\ul{TapLab significantly reduces redundant computations of the state-of-the-art fast semantic image segmentation models, running 3 to 10 times faster with controllable accuracy degradation.} The experimental results show that TapLab achieves 70.6\% mIoU on the Cityscapes dataset at 99.8 FPS with a single GPU card for the 1024$\times$2048 videos. A high-speed version even reaches the speed of 160+ FPS. Codes will be available soon at \url{https://github.com/Sixkplus/TapLab}. 
\end{abstract}

}

\maketitle

\IEEEdisplaynontitleabstractindextext

%
\IEEEpeerreviewmaketitle

\IEEEraisesectionheading{\section{Introduction}\label{sec:intro}}

\IEEEPARstart{S}{e}mantic segmentation  
is typically cast as pixelwise classification on unstructured images or videos. 
Being effective in feature representation and   
discriminative learning, convolutional neural networks (CNNs) \cite{lecun1989backpropagation} have been working as a popular and powerful tool for semantic segmentation. 
With the advent of high-resolution (e.g., 1080p and 4K) videos, conventional CNN-based segmentation approaches usually impose high computational and memory costs which hinder real-time applications. 
Fast semantic video segmentation with high accuracy is an urgent demand for high-resolution vision applications.

\begin{figure}[tb]
\begin{center}
   \includegraphics[width=1.0\linewidth]{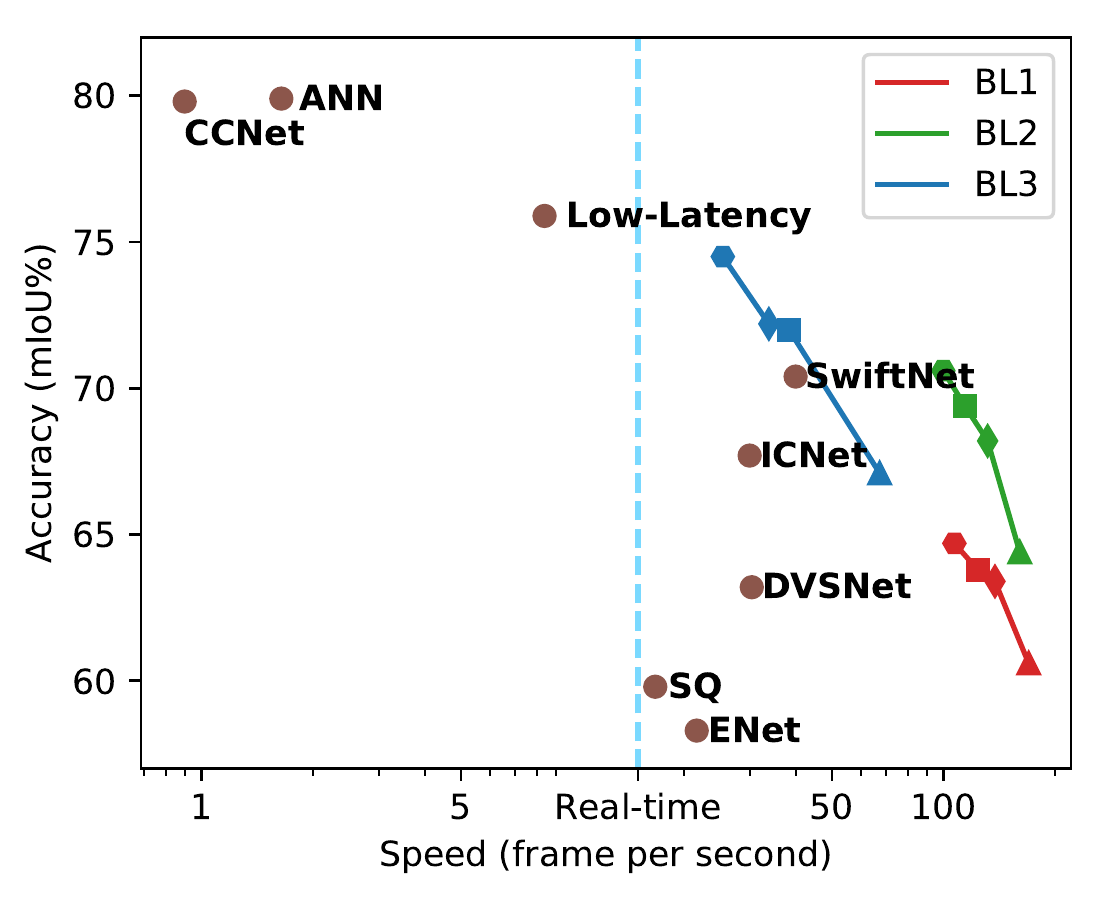}
\end{center}
	\vspace{-2em}
   \caption{Comparison of different approaches for semantic video segmentation at a resolution of $1024\times2048$ \ul{on the Cityscapes dataset}. The brown dots denote existing methods. The red, green, and blue marks denote results with our first, second, and third baseline model respectively. The triangles denote the results with the FFW module. The diamonds denote the results with FFW and RGC modules. The squares denote the results with FFW and RGFS modules. The hexagons denote the results with FFW, RGC, and RGFS modules. The real-time reference line is set at 15 FPS. Our approach gains a huge advantage in terms of inference time and achieves comparable accuracy compared with other real-time methods. Notice that the horizontal axis is logarithmic.
   }
\label{fig:acc-eff}
\end{figure}
A typical way of semantic \emph{video} segmentation treats a video clip as a sequence of individual frames, relying on a network for semantic \emph{image} segmentation~\cite{long2015fully,chen2018deeplab,chen2018encoder} to perform segmentation in a frame-by-frame fashion.
To meet the real-time demand, such segmentation approaches usually trade off lower accuracy for faster speed by reducing the input scale or designing a lightweight network~\cite{wu2017real,badrinarayanan2017segnet,paszke2016enet, zhao2018icnet, yu2018bisenet, li2019dfanet, orsic2019defense}.
However, these segmentation approaches ignore the temporal continuity of videos,
thereby leading to the redundant computational burden across frames~\cite{siam2018comparative}.

In light of the above issue, a number of segmentation approaches introduce an extra
temporal feature extraction module to model the continuity of neighboring frames 
by 3D CNNs~\cite{carreira2017quo,tran2015learning}, 
RNNs~\cite{donahue2015long,yue2015beyond}, or optical flow estimation~\cite{zach2007duality,sun2018pwc}.
Based on temporal features, only keyframes, which account for a small percentage of all the frames, require full segmentation, while the other frames undergo cross-frame feature propagation or label propagation. Although the above segmentation pipelines speed up their inference phase, they usually have heavy costs incurred by temporal feature extraction, e.g., optical flow estimation, which is itself a bottleneck for real-time performance.  

In general, videos are compressed data in the form of computer files and network streaming. Videos in the compressed domain already contain a rich body of motion information such as motion vectors~($\mathbf{Mv}$) and residuals~($\mathbf{Res}$). 
Recently, these compressed-domain features have been tapped in video tasks to avoid the cost incurred by video decoding and the aforementioned temporal feature extraction.
Despite the fact that motion vectors are noisier (superpixel-level instead of pixel-level), such video-level tasks as video classification \cite{chadha2017compressed}, action recognition \cite{Wu_2018_CVPR} and vehicle counting \cite{liu2016highway} can tolerate the noise.
On the contrary, it takes special efforts to apply coarse-grained compressed-domain features to semantic segmentation, a pixel-level task, to achieve high accuracy.

Inspired by the above observations, we propose a novel real-time semantic video segmentation framework, named TapLab, utilizing motion information from the compressed domain for efficiency.
The framework consists of a semantic image segmentation network and three plug-and-play modules tailored for semantic video segmentation. Specifically, we design a fast feature warping~(FFW) module that exploits motion vectors for feature and label propagation across consecutive frames. The experimental results show that this module reduces the inference time by a wide margin.
To address the noise problem introduced by motion vectors, we design a residual-guided correction~(RGC) module, which adaptively selects the most inconsistent region for further refinement, and furthermore, we design a residual-guided frame selection (RGFS) module to determine the hard-to-warp frames and do segmentation instead of warping for them. 
The experiments demonstrate these two modules are able to refine the coarse segmentation results and improve the model's robustness. 
As a result, TapLab significantly reduces redundant computations of the semantic image segmentation models, running 3 to 10 times faster with controllable accuracy degradation, as illustrated in Fig.~\ref{fig:acc-eff}. Also, we show that our modules are generic to networks for semantic image segmentation.

In summary, the contributions of this work are two-fold.
    First, we propose a novel real-time semantic video segmentation framework that taps into the encoded features that already exist in videos. In addition to a CNN for semantic segmentation, the proposed framework includes three modules: a fast feature warping module to utilize the temporal continuity in videos, a residual-guided correction module to refine local regions, and a residual-guided frame selection module to select the hard-to-warp frames for segmentation.
    Second, the experiments demonstrate our modules are generic to a variety of segmentation networks and the framework achieves around 3 to 10 $\times$ speed-up against the semantic image segmentation networks with controllable accuracy degradation. On the Cityscapes~\cite{cordts2016cityscapes} dataset, 
	TapLab obtains the results of 70.6\% mIoU with on 1024$\times$2048 input at 99.8 FPS with a single GPU card. A high-speed version of TapLab achieves an FPS of 160.4 with 64.4\% mIoU. 

\section{Related Work}
\label{sec:related}
\subsection{Fast Image Segmentation}
Driven by the development of deep CNNs, semantic segmentation approaches~\cite{zhao2017pyramid,chen2018deeplab,chen2018encoder, cfnet, ccnet, danet, ann} based on FCN~\cite{long2015fully} have achieved surprisingly high accuracy. Recently, more works have changed the focus onto efficiency~\cite{siam2018comparative}. Early works~\cite{paszke2016enet,badrinarayanan2017segnet,wu2017real} either downsample the inputs or prune the channels of their networks.
ICNet~\cite{zhao2018icnet} and BiSeNet~\cite{yu2018bisenet} propose multi-path strategies in which a deeper path with faster downsampling is designed to extract context features while a shallow path with original scale to preserve local details. Moreover, efficient fusion modules are assigned to combine features from different paths. More recently, SwiftNet \cite{orsic2019defense} and DFANet \cite{li2019dfanet} propose lightweight networks with pyramid fusion or aggregation for features.
However, these methods deal with images or consider a video as individual frames. Thus, they are incapable of leveraging the temporal continuity of videos.


\subsection{Semantic Video Segmentation}
Methods dealing with video tasks tend to capitalize on temporal continuity in videos and thus to extract various kinds of temporal features, among which optical flow is the most commonly used one~\cite{li2018low,gadde2017semantic,zhu2017deep,xu2018dynamic}. 
FlowNet~\cite{dosovitskiy2015flownet} and FlowNet 2.0~\cite{ilg2017flownet} estimate optical flow fields based on DCNNs and are able to run at high speed, followed by 
many flow-based segmentation strategies~\cite{gadde2017semantic,zhu2017deep,xu2018dynamic}.
Gadde et al.~\cite{gadde2017semantic} employ optical flow to warp features from different layers for feature quality enhancement. Zhu et al.~\cite{zhu2017deep} and Xu et al.~\cite{xu2018dynamic} utilize the efficiency of FlowNet to propagate results of keyframes for model acceleration. However, due to the extra time consumed by flow estimation, these models perform on par with fast per-frame models.

The aforementioned flow-based methods rely heavily on keyframes scheduling strategies.
 Zhu et al.~\cite{zhu2017deep} preset a fixed interval to determine keyframes. 
Adaptive scheduling strategies, e.g.,~\mbox{\cite{ilg2017flownet}} and~\mbox{\cite{xu2018dynamic}}, determine keyframes according to confidence scores calculated by a lightweight CNN branch.
In addition to dynamic keyframe selection, Xu et al.~\mbox{\cite{xu2018dynamic}} also divide a single frame into small regions and heuristically selects less confident ones to pass through the whole segmentation network. In the area of video object detection, Zhu et al.~\mbox{\cite{zhu2018towards}} also propose to warp features across adjacent frames and learn to select key regions/frames to perform refinement.

To our knowledge, TapLab is the first work to utilize the existing encoded features \textit{residual maps} to select keyframes and key regions, making the selection procedure training-free, generic to various datasets, and extremely fast.

\subsection{Compressed-Domain Video Analysis} 
Recently, features from compressed data have been utilized in vision tasks such as video classification~\cite{Chadha:2019hu,chadha2017compressed}, vehicle counting~\cite{Wang:2017ja,liu2016highway}, action recognition~\cite{Shou_2019_CVPR, Wu_2018_CVPR}, etc. Despite the fact that compressed-domain features are noisier than pixel-domain, these video-level tasks can tolerate the noise. On the contrary, it takes special efforts to apply noisy compressed-domain features to
semantic segmentation, a pixel-level task, to achieve high accuracy.
More recently, Jain et al.~\cite{jain2018fast} design a bidirectional feature warping module with motion vectors for semantic segmentation. However, the bidirectional feature warping design produces latency and does not solve the problem of precision-degrading caused by motion vectors. 


\section{Methods}
\label{sec:methods}
\begin{table}[tb]
\centering
\caption{Notations}
\label{table:notation}
	\vspace{-1em}
  \begin{tabu} to 0.49\textwidth {X[c]X[2]}
\toprule
$\mathbf{I}$ & original RGB frame\\
$\mathbf{I}_s$ & shifted frame\\
$\mathbf{Mv}$ & motion vector\\
$\mathbf{Res}$ & residual map\\
$\mathbf{R}_i$ & the $i^{th}$ region\\
$\mathcal{F}$ & feature map\\

\midrule
$p$ & pixel or element index \\
$t$ & current frame index \\

\midrule
FFW& fast feature warping\\
RGC& residual-guided correction\\
RGFS& residual-guided frame selection\\
$\phi$& segmentation CNN  \\
\midrule
	  $\mathrm{THR_{RGC}}$ & threshold for the RGC module \\
	  $\mathrm{THR_{RGFS}}$ & threshold for the RGFS module \\

\bottomrule
\end{tabu}
\end{table}

	
In this section, we present details of our framework TapLab. We first introduce the basics of compressed video. Next, we describe our video segmentation framework consisting of a segmentation model and three plug-and-play modules tailored for semantic video segmentation, i.e., a \textit{fast feature warping}~(FFW) module, a \textit{residual-guided correction}~(RGC) module, and a \textit{residual-guided frame selection}~(RGFS) module. Finally, we present the implementation details. For convenience, Table~\ref{table:notation} summarizes the notations.

\subsection{Basics of Compressed Video}

In general, an encoded video stream consists of groups of pictures~(GOP). 
A GOP can contain three types of frames: I-frame, P-frame, and B-frame. An I-frame, a coded frame independent of all other frames, marks the beginning of a GOP. A P-frame is predicted from its previous I-frame or P-frame and a B-frame is predicted from its previous and next I-frame or P-frame. A typical sequence of a GOP can be IPPBPPPPPPBP.

\begin{figure}[tb]
\begin{center}
   \includegraphics[width=0.999\linewidth]{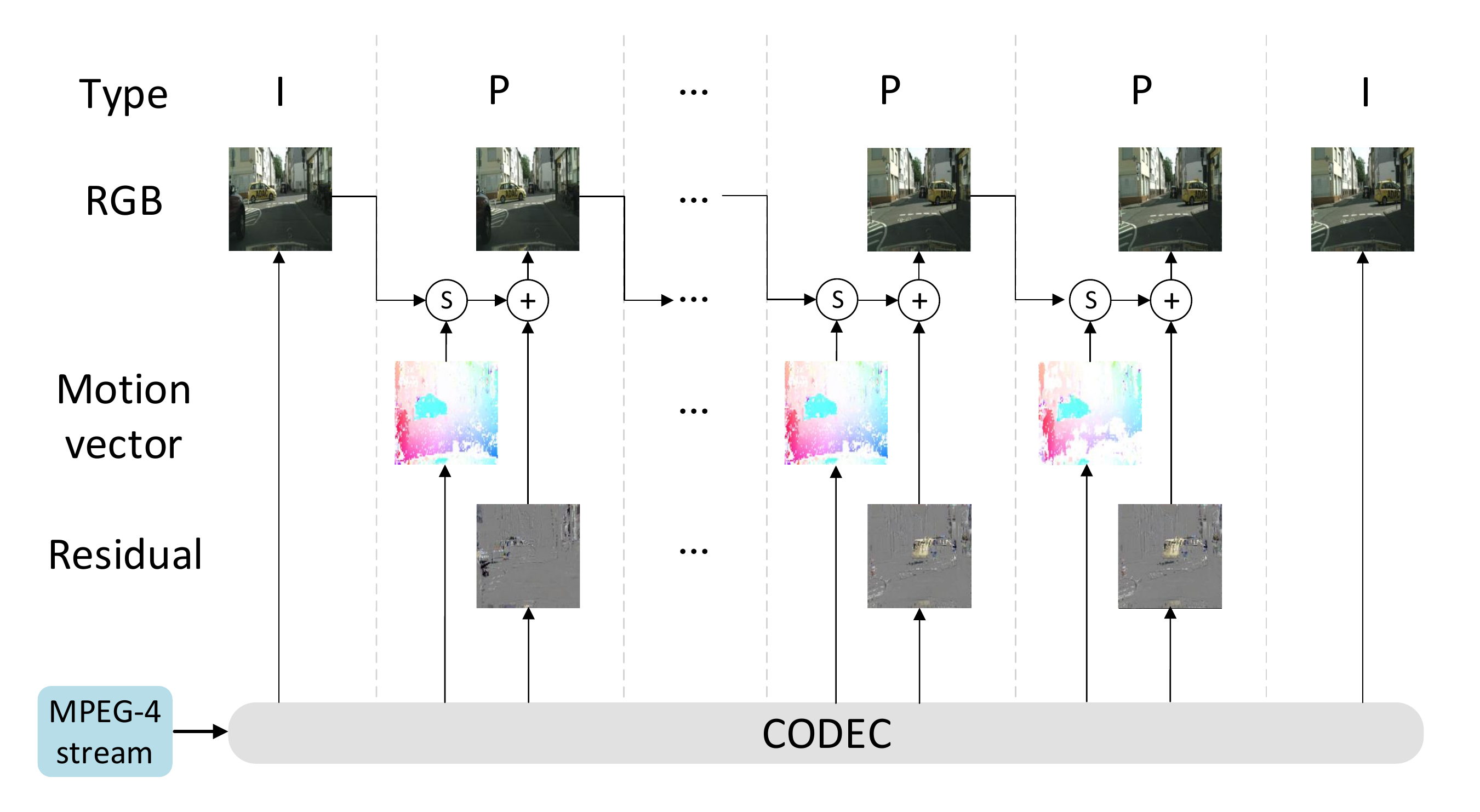}
\end{center}
	\vspace{-2em}
   \caption{Illustration of decoding process. An MPEG-4 stream consists of I-frames and P-frames. An I-frame is independently encoded, while a P-frame is generated from motion compensation  with motion vectors and residuals. ``S'' stands for the shifting of pixels from a reference frame to a predicted frame and  ``+'' for element-wise addition.
}
\label{fig:codec}
\end{figure}

We use videos encoded by MPEG-4 Part 2 (Simple Profile)~\cite{le1991mpeg}, following recent work of \cite{Wu_2018_CVPR} and \cite{Shou_2019_CVPR} in the compressed domain. A default GOP in this standard contains an I-frame followed by 11 P-frames (no B-frame). In the compressed domain, as shown in Fig.~\ref{fig:codec}, three types of data are readily available: (1) I-frames, the beginning encoded frames of each GOP, (2) motion vectors~($\mathbf{Mv}$), the displacement of a P-frame from the previous frame, either an I-frame or a P-frame, and (3) residuals~($\mathbf{Res}$), the difference between a P-frame and its referenced motion-compensated frame. It is worth noting that motions vectors and residuals are encoded in many popular codecs, such as MPEG, H.264, H.265. Without loss of generality, we use MPEG-4 in our experiments. The framework can be easily generalized to other codec standards.

\label{ssec:basics}
Features in the compressed domain are coarse-grained. During compression, each frame is typically divided into 16x16 macroblocks and motion vectors represent the displacement of the macroblocks. As a result, motion vectors have a much lower resolution. Although previous works~\cite{Wang:2017ja,liu2016highway,Chadha:2019hu,chadha2017compressed,Shou_2019_CVPR,Wu_2018_CVPR} show their effectiveness in video-level classification problems, it is impractical to directly apply them to semantic segmentation, which requires pixel-level predictions. Thus, we design the following framework. 


\subsection{Framework}


\begin{figure*}[tb]
\begin{center}
   \includegraphics[width=0.99\linewidth]{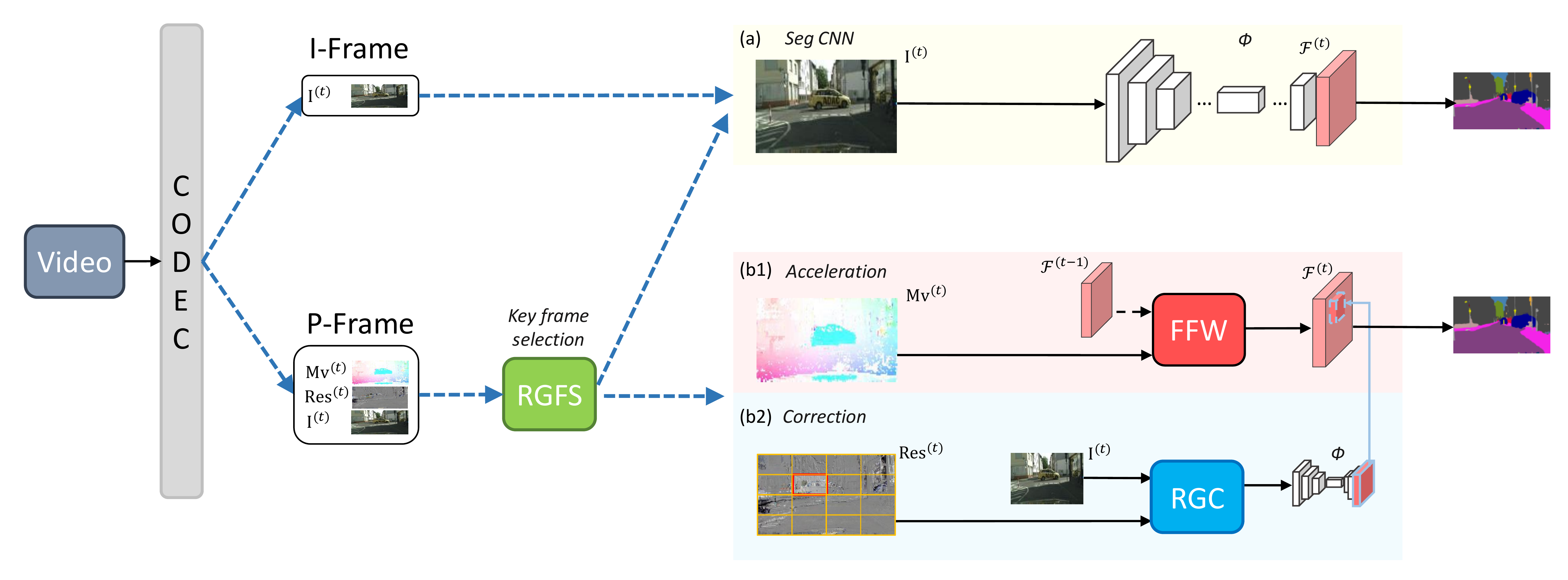}
\end{center}
	\vspace{-2em}
   \caption{An overview of the proposed semantic video segmentation framework. 
   All I-frames are directly sent to the segmentation networks. For P-frames, the RGFS module takes the residual maps as input and decides whether the current frame should be sent to (a) or (b1, b2). (a)~Baseline segmentation network. It takes the whole frame as input and outputs the result feature maps. (b1)~Acceleration. The fast feature warping (FFW) module takes as input motion vectors and feature maps from the previous frame. It speeds up the segmentation by a wide margin. (b2)~Correction. The residual-guided correction~(RGC) module selects a region based on the residual map and performs local segmentation. ``$\phi$'' denotes the baseline segmentation CNN. The blue arrows represent the decision-related procedure. 
}

\label{fig:FVSCD}
\end{figure*}

As illustrated in Fig.~\ref{fig:FVSCD}, our segmentation framework consists of a CNN for semantic image segmentation and three modules tailored for semantic video segmentation based on compressed-domain features. The CNN (baseline model) could be any network for semantic image segmentation, and we choose three commonly used architectures. 
As for the modules, we concentrate on speeding up the segmentation for P-frames. First, to accelerate the segmentation process, we design the fast feature warping~(FFW) module to propagate spatial features based on motion vectors.
Second, we design the residual-guided correction~(RGC) module to refine local segmentation. RGC selects the ``worst'' region of a current frame and performs fine segmentation for this region.
Third, we design the residual-guided frame selection~(RGFS) module to refine a small portion of P-frames. RGFS selects the ``hard-to-warp'' P-frames and sends them into the segmentation CNN adaptively. 

In addition to the components, Fig.~\ref{fig:FVSCD} shows the complete data flow of the proposed framework and the connections among different modules. After decoding, all the I-frames are directly sent to the segmentation network. As for P-frames, RGFS selects the P-frames needed to be sent to the CNN. The rest P-frames are processed with FFW and RGC.

\begin{figure*}[tb]
\begin{center}
   \includegraphics[width=0.99\linewidth]{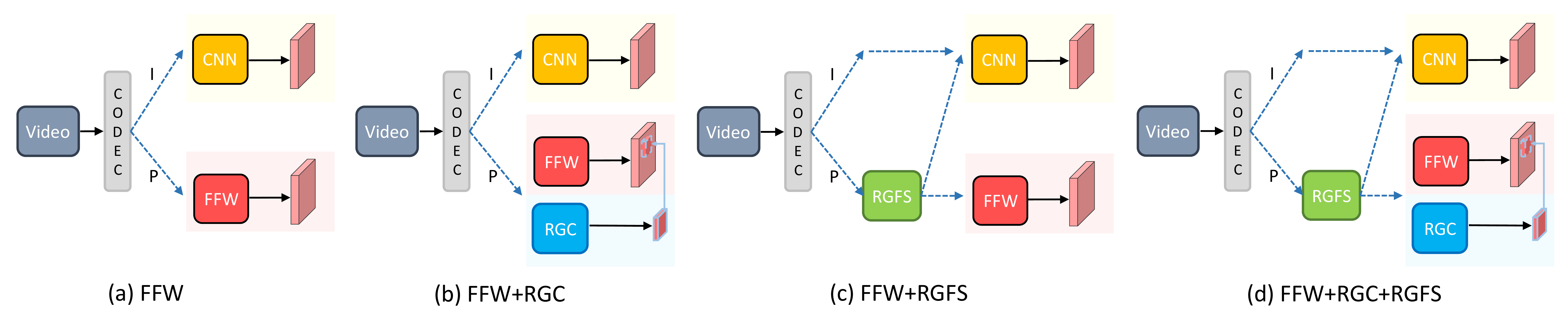}
\end{center}
	\vspace{-2em}
   \caption{Different combinations of the proposed modules. For simplicity, the inputs of the modules are omitted.
}

\label{fig:plugin}
\end{figure*}

It is worth noting that our framework has different versions. Based on the core module FFW, the RGC module and the RGFS module can be treated as plug-ins and be added to or removed from the whole framework easily. Fig.~\ref{fig:FVSCD} only shows the most complicated case~(FFW+RGC+RGFS). More combinations~(FFW, FFW+RGC, FFW+RGFS) of modules are shown in Fig.~\ref{fig:plugin}. The plug-and-play design gives more choices to strike a balance between accuracy and speed according to the actual requirements.

We describe the details of each component below.

\subsubsection{Baseline Segmentation Models}
\label{sec:baseline}
We start building TapLab from choosing the semantic image segmentation models. To demonstrate the effectiveness and genericity of our modules, we use three different commonly used segmentation CNN architectures as the baseline models following the recent works of~ICNet\cite{zhao2018icnet}, U-Net\cite{ronneberger2015u}, and~PSPNet~\cite{zhao2017pyramid}. During the process of semantic video segmentation, each I-frame is fed into a segmentation model, denoted by $\phi$, and each P-frame can be speeded up by using compressed-domain features. The $\phi$  could also take P-frames for refinement. Next, we will describe our modules for speeding up segmentation for P-frames.
\subsubsection{Fast Feature Warping}
	Considering the transformation consistency of input images and the corresponding output labels in semantic segmentation, we design the fast feature warping~(FFW) module. This module takes in the previous feature maps $\mathcal{F}^{(t-1)}$ and the current motion vectors $\mathbf{Mv}^{(t)}$ and outputs the current feature maps $\mathcal{F}^{(t)}$.  The warping in the feature domain is equivalent to shifting in the pixel domain.
 Thus, $\mathcal{F}^{(t)}$ is defined as 
\begin{equation}
\begin{split}
	\mathcal{F}^{(t)}[p]& = \mathrm{FFW}(\mathbf{Mv}^{(t)}, \mathcal{F}^{(t-1)})[p]\\
	&= \mathcal{F}^{(t-1)}[p - \mathbf{Mv}^{(t)}[p]],
\end{split}
\label{eq:warp}			
\end{equation}
where $p = (x,y) \in H\times W$ represents the ``pixel'' index in the feature maps. According to Equation~\eqref{eq:warp}, there are just simple shifting operations during FFW, making this procedure extremely fast.

To make the procedure even faster, we could use longer GOPs.
Given the GOP number $g$ and inference time $T_I$, $T_P$ for I-frames and P-frames respectively, the overall inference time is defined by 
\begin{equation}
\begin{split}
	T_{avg} = \frac1{g}\cdot T_I + (1-\frac1g) \cdot T_P,
\end{split}
\label{eq:time}			
\end{equation}
which indicates that if $T_P \ll T_I $, larger $g$ makes for higher speed. We study the influence of GOP number on accuracy in Sec. \ref{ssub:ffw_eval}.

\begin{figure}[tb]
\begin{center}
   \includegraphics[width=0.999\linewidth]{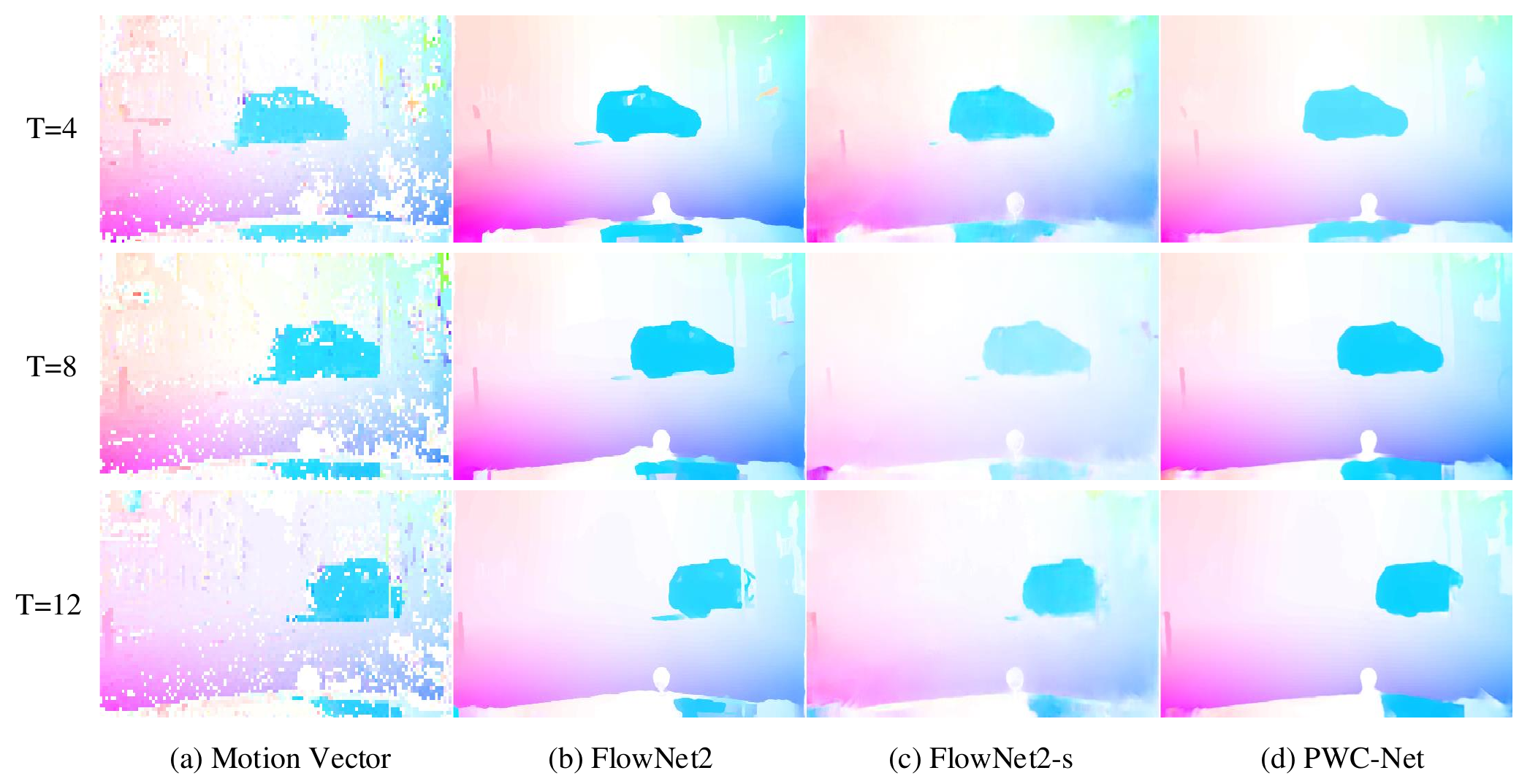}
\end{center}
	\vspace{-2em}
   \caption{Visualization of motion vector and optical flow. Following the work of Wu et al.~\mbox{\cite{Wu_2018_CVPR}}, we convert the 2D motion values into 3D HSV values, where hue and saturation refer to the direction~(angle) and magnitude of the motion respectively. In general, both motion vectors and optical flow fields can correctly represent most kinds of movements. Optical flow fields contain more details at the original resolution. }
\label{fig:mvof}
\end{figure}

Actually, optical flow-based methods~\cite{xu2018dynamic,zhu2017deep} also use warping for speeding-up.
We take motion vectors rather than  optical flows as the input of the warping module for the following considerations.
First, the use of motion vectors makes the framework faster. Motion vectors are compressed-domain features that already exist in videos. They can be accessed with ease while optical flow estimation takes considerable extra time.
Second, motion vectors, albeit coarse-grained~(shown in Fig.~\ref{fig:mvof}(a)), fit the modern semantic segmentation CNN models and perform on a par with optical flow estimation in terms of segmentation accuracy, as shown in Table~\ref{table:mv}. Motion vectors store the motion information of small blocks~(usually areas of $16\times 16$ pixels), while optical flow algorithms calculate the motion information of all the pixels~(shown in Fig.~\ref{fig:mvof} (b, c, d)).  Nevertheless, most segmentation CNNs utilize pooling layers and convolution layers with strides to obtain a larger receptive field and get more context information, resulting in a smaller shape of feature maps~(usually $1/16$ or $1/8$ of the input image). Therefore, the block-level motion information of motion vectors is sufficient for feature warping. Also, experimental results demonstrate that the accuracy of flow information is not directly related to the segmentation accuracy. 
Fig.~\ref{fig:mvof} shows the motion vector and the optical flow of a sample frame.

Despite the high efficiency, warping-based segmentation models display weak robustness, since neither motion vectors nor optical flow fields can present all kinds of movements, e.g., the appearance of new objects. Hence, previous works~\mbox{\cite{xu2018dynamic, li2018low}} adaptively select keyframes for fine-segmentation. We rethink this problem from the perspective of codec principles and design the following RGC and RGFS modules.

\subsubsection{Residual-Guided Correction}
\label{sec:rgc}

\begin{figure}[tb]
\begin{center}
   \includegraphics[width=0.999\linewidth]{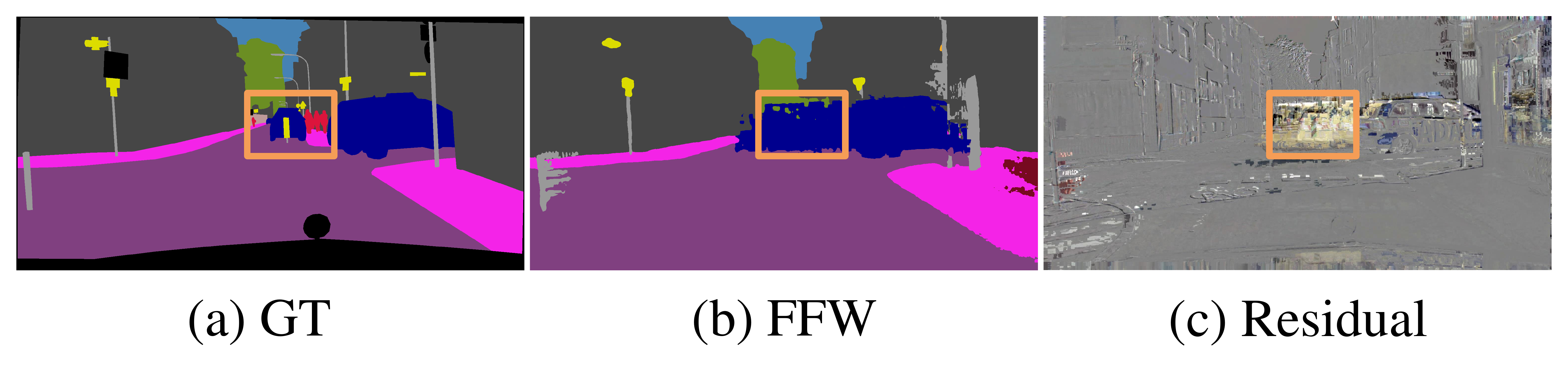}
\end{center}
	\vspace{-2em}
   \caption{An example of residual map. The region with high values in the residual map~(c) corresponds to the region where the segmentation result of FFW~(b) is poor.}
\label{fig:residual}
\end{figure}

In modern video coding algorithms, to handle the inevitable differences between the shifted image~$\mathbf{I}_s$ and the original one~$\mathbf{I}$, element-wise residual maps for compensation are introduced~\cite{sofokleous2005h}. Inspired by this operation, we propose the residual-guided correction~(RGC) module. This module takes residual maps as input and adaptively selects one region for fine-segmentation. The absolute value in residual maps at a certain point $|\mathbf{Res}[p]|$ describes the difference between~\mbox{$\mathbf{I}_s[p]$} and~\mbox{$\mathbf{I}[p]$}. Thus, a region~\mbox{$\mathbf{R}_i = H_i \times W_i \times C$} with higher magnitudes in~\mbox{$\mathbf{Res}$} indicates we have lower confidence for its warped feature map~$\mathcal{F}[\mathbf{R}_i]$~(e.g. the example in Fig.~\mbox{\ref{fig:residual}}). We divide the whole frame by grids and select the one with the highest magnitude in the corresponding residual map. Accordingly, the selection policy of RGC is defined as 
\begin{equation}
	\mathrm{RGC}(\mathbf{Res}^{(t)})= \mathop{\arg \max}_{\mathbf{R}_i} \sum_{p \in \mathbf{R}_i} I(\ |\mathbf{Res}^{(t)}[p]| > \mathrm{THR_{RGC}}\ ),
\label{eq:bcm}			
\end{equation}
where $i$ is the region index and $I(*)$ is the indicator variable which takes value 1 if $(*)$ is true and 0 otherwise, and $\mathrm{THR_{RGC}}$ is a threshold to avoid noise. 
After selection, the chosen region is sent to the segmentation CNN for refinement.

	Compared with commonly used region of interest~(ROI) selection algorithms such as SS~\mbox{\cite{uijlings2013selective}} and RPN~\mbox{\cite{ren2015faster}}, our training-free RGC is faster, simpler and more intuitive.

\subsubsection{Residual-Guided Frame Selection}
In addition to refining selected spatial regions, we capitalize on residual maps to adaptively select keyframes that are ``hard-to-warp''. For each P-frame, we calculate the frame-level residual score as 
\begin{equation}
	\mathrm{RGFS}(\mathbf{Res}^{(t)})= \sum_{p \in \mathbf{Res}^{(t)}} |\mathbf{Res}^{(t)}[p]|.
\label{eq:rgfs}			
\end{equation}
Similar to the analysis in Section~\ref{sec:rgc}, the summation of absolute values in a residual map indicates the quality of the corresponding motion vector. The higher the residual score, the higher probability that the warped result is untrustworthy. In such situations, the corresponding frames are sent into the CNN for fine-segmentation.

We set a threshold $\mathrm{THR_{RGFS}}$ for the RGFS module to select the ``hard-to-warp'' frames. If $\mathrm{RGFS}(\mathbf{Res}^{(t)})>\mathrm{THR_{RGFS}}$, the current P-frame is treated as a keyframe. Higher $\mathrm{THR_{RGFS}}$ indicates that the module is less sensitive to the noise of MV, and the average inference speed becomes faster due to fewer keyframes. As a trade-off, the accuracy would decrease.

Compared with~\cite{xu2018dynamic,li2018low} which apply dynamic keyframe selection by adding a CNN branch to predict the confidence score, RGFS is simpler and faster. Moreover, the residual-guided modules are intuitive since residual maps are meant to offer motion compensation.

\begin{algorithm}[tb]
\setstretch{1.25}
\caption{Inference Procedure} 
\label{alg:Framwork} 
\begin{algorithmic}[1] 
\REQUIRE~\\ 
The compressed video stream\ \ $\mathbf{V}$;

\STATE  $\mathbf{for}$ t = 1 $\mathbf{to}$\ $|\mathbf{V}|\ \  \mathbf{do}$

\STATE \ \ \ \   $\mathbf{if}$\ \ $t^{th}$\  frame\ is\ I-type\ \ $\mathbf{then}$

\STATE \ \ \ \ \ \ \ \ decode $\mathbf{I}^{(t)}$
\STATE \ \ \ \ \ \ \ \ $\mathcal{F}^{(t)} = \phi(\mathbf{I}^{(t)})$ 
\STATE \ \ \ \   $\mathbf{else\ \ do} $
\STATE \ \ \ \ \ \ \ \ decode $\mathbf{Mv}^{(t)}$, $\mathbf{Res}^{(t)}$, $\mathbf{I}^{(t)}$
	\STATE \ \ \ \ \ \ \ \ $\mathbf{if}\ \  \mathrm{RGFS}(\mathbf{Res}^{(t)})$\ \  $> \mathrm{THR_{RGFS}}$\ \ $\mathbf{then}$
\STATE \ \ \ \ \ \ \ \ \ \ \ \ $\mathcal{F}^{(t)} = \phi(\mathbf{I}^{(t)})$
\STATE \ \ \ \ \ \ \ \ $\mathbf{else\ \ do} $
\STATE \ \ \ \ \ \ \ \ \ \ \ \ $\mathcal{F}^{(t)} = \mathrm{FFW}(\mathbf{Mv}^{(t)}, \mathcal{F}^{(t-1)})$
\STATE \ \ \ \ \ \ \ \ \ \ \ \ $\mathbf{R}_i^{(t)} = \mathrm{RGC}(\mathbf{Res}^{(t)})$
\STATE \ \ \ \ \ \ \ \ \ \ \ \ $\mathcal{F}^{(t)}[\mathbf{R}_i^{(t)}] = \phi(\mathbf{I}^{(t)}[\mathbf{R}_i^{(t)}])$

\STATE \ \ \ \   $\mathbf{Output:}$ current segmentation result $\mathcal{F}^{(t)}$
\end{algorithmic}
\end{algorithm}

\subsection{Implementation Details}
Here are the implementation details of our loss function and inference algorithm.

\subsubsection{Loss Function}

To train the baseline segmentation CNNs, we follow the previous works and use the softmax cross-entropy loss defined as 

\begin{equation}
        \mathcal{L} = -\sum_{x = 1}^{H}\sum_{y = 1}^{W}\log\frac{e^{\mathcal{F}(x,y,c_g)}}{\sum_{c=1}^C e^{\mathcal{F}(x,y,c)}},
\label{eq:loss}
\end{equation}
where $c_g$ is the ground truth class.

\subsubsection{Inference Algorithm}
	The overall inference procedure is summarized in Algorithm~\ref{alg:Framwork}.

        Considering the implementation complexity, we only encode I-frames and P-frames during compression. Note that the weights of the CNN in the RGC module are the same as those of the per-frame segmentation model. 
	
	For the RGC module, the threshold is universal for different datasets. Empirically, $\mathrm{THR_{RGC}} \in \{10,20,30,40\}$ leads to similar performance.
For the RGFS module, we choose $\mathrm{THR_{RGFS}}$ such that about $10\%$ P-frames are selected as keyframes. This parameter can be adjusted to balance speed and accuracy. We choose this threshold on the training set of different datasets.

\section{Experimental Evaluation}
\label{sec:eval}



In this section, we evaluate TapLab on high-resolution videos. We first briefly introduce the experimental environment. Then we perform ablation studies to validate the effectiveness of each module. Finally, we perform a thorough comparison of our model with the state-of-the-art fast segmentation models in terms of both accuracy and speed. 

\subsection{Experimental Environment}
\subsubsection{Datasets}

\begin{table}[b]
	\caption{Performance of Baseline Segmentation Models \ul{on Cityscapes}} \label{table:baseline}
	\vspace{-1em}
  \begin{tabu} to 0.49\textwidth {X[2]X[2]X[c]X[c]}
\toprule
Model&Backbone&mIoU&FPS\\
\midrule
BL1&ResNet-50$\dagger$ &67.3&33.2 \\
BL2&MobileNet&73.6&29.3 \\
BL3&ResNet-101&77.3&7.2 \\
\bottomrule
\end{tabu}
  
  \medskip
  \emph{\footnotesize  All the backbones are not pre-trained. The results are evaluated on the validation set. ``$\dagger$'': a modified lightweight version.}

\end{table}

There exist many commonly used datasets for the semantic segmentation task, such as Cityscapes~\cite{cordts2016cityscapes}, CamVid~\cite{brostow2009semantic}, COCO-Stuff~\cite{caesar2018coco}, ADE20K~\cite{zhou2017scene}, and so on. Considering the demand for high-resolution input and the requirement that there should be image sequences to form video clips, we choose to perform training and validation mainly on Cityscapes, a dataset for semantic understanding of urban street scenes. 
It contains 11 background categories and 8 foreground categories. 
The 5000 finely annotated images are split into training, validation and testing sets with 2975, 500, and 1525 images respectively.
        Each of these images is actually the $20^{th}$ frame of a 30-frame video clip. All the frames have a resolution of 1024$\times$2048.

In addition to the main ablations on Cityscapes, we also provide qualitative and quantitative results on CamVid.


\subsubsection{Protocol}
In our experiments, we choose MPEG-4 Part 2 (Simple Profile)~\cite{le1991mpeg} as the compression standard where the B-frame rate is 0. The Group of Pictures~(GOP), which determines the interval for two adjacent I-frames, defaults to 12. 

As for the details of our modules, we choose regions at a resolution 512$\times$512 and the stride along each axis is~256 for our RGC module. The noise threshold $\mathrm{THR_{RGC}}$ for compensation map judgment is set to~30, and the threshold $\mathrm{THR_{RGFS}}$ for the RGFS module is set to $3.6 \times 10^7$.

We evaluate the performance on the validation set. We randomly choose the interval between the starting frame and the test frame since only one frame of the 30-frame video clip is annotated. No testing augmentation like multi-scale or multi-crop is employed. We evaluate the speed and accuracy on images at a resolution of $1024\times 2048$ using only the single-scale model. The accuracy is measured by mean Intersection-over-Union (mIoU). All the experiments are performed on a server with an Intel Core i7-6800K CPU and a single NVIDIA GeForce GTX 1080 Ti GPU card. We use TensorFlow~\cite{abadi2016tensorflow} to build the CNNs.

\subsection{Ablation Study}

	

\begin{table}[b]
	\caption{Performance of Feature Propagation Methods \ul{on Cityscapes}}
	\vspace{-1em}
  \begin{tabularx}{0.49\textwidth}{Xcccc}
\toprule
Model&$T_{flow}~(ms)$&$T_{warp}~(ms)$&$T_{total}~(ms)$&mIoU\\
\midrule
BL1+ITP&-&20&20&36.1\\
BL1+flow2&67&3.7&70.7&61.1\\
BL1+flow2C&32&3.7&35.7&60.2\\
BL1+flow2S&23&3.7&26.7&59.8\\
BL1+PWC&19&3.7&22.7&60.4\\
BL1+FFW&-&3.7&3.7&60.6\\
\midrule
BL2+ITP&-&20&20&42.1\\
BL2+flow2&67&3.7&70.7&62.1\\
BL2+PWC&19&3.7&22.7&61.7\\
BL2+FFW&-&3.7&3.7&64.4\\
\midrule
BL3+ITP&-&20&20&43.1\\
BL3+flow2&67&3.7&70.7&67.5\\
BL3+PWC&19&3.7&22.7&66.8\\
BL3+FFW&-&3.7&3.7&67.1\\
\bottomrule
\end{tabularx}

\medskip
  \emph{\footnotesize $T_{flow}$: the time for extracting optical flows. $T_{warp}$: the time for warping or interpolation, $T_{total}$: the total running time for warping~(interpolation). ``ITP'': interpolation method. ``flow2'': FlowNet~2.0~\cite{ilg2017flownet} for optical flow estimation. ``PWC'': PWC-Net~\cite{sun2018pwc}. ``FFW'': the fast feature warping module.}
\label{table:mv}
\end{table}
\begin{figure}[tb]
\begin{center}
   \includegraphics[width=0.99\linewidth]{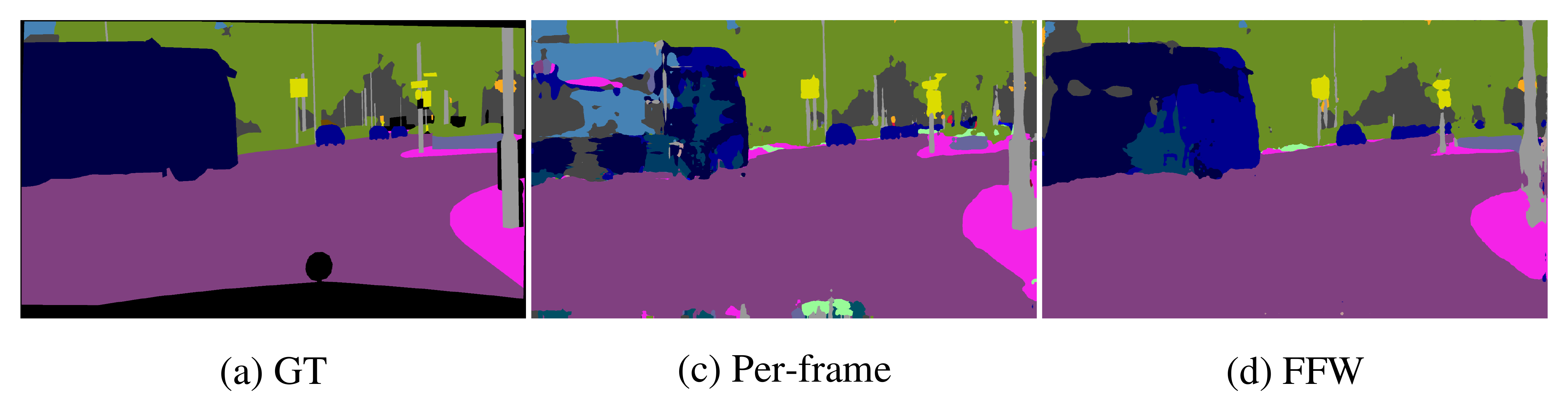}
\end{center}
	\vspace{-2em}
   \caption{A moving vehicle across the camera view. In this case, the result with the FFW module is more accurate.}

\label{fig:ffwbt}
\end{figure}

\begin{figure}[tb]
\begin{center}
   \includegraphics[width=0.99\linewidth]{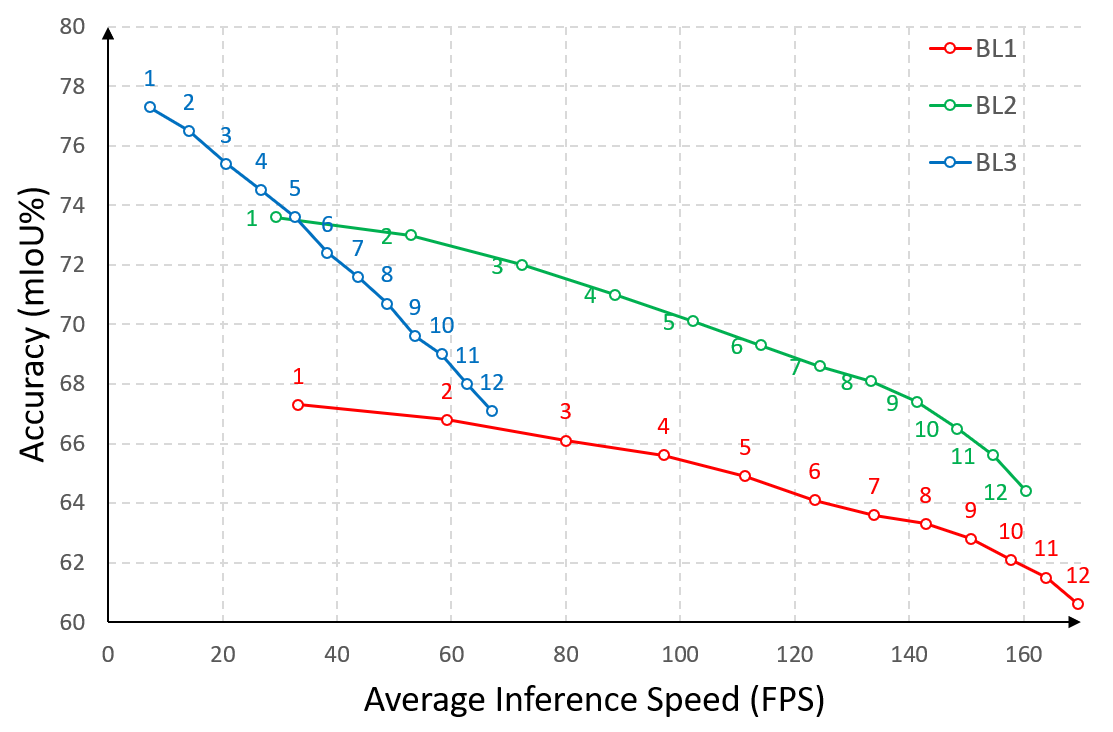}
\end{center}
	\vspace{-2em}
   \caption{Accuracy~(mIoU) versus speed~(FPS) under different GOP configurations. The number above each point indicates the GOP number.}
\label{fig:gop}
\end{figure}
\subsubsection{Baseline}
\label{sssec:baseline}
We start building our semantic video segmentation framework from the implementation of per-frame segmentation CNN models. As described in Section~\ref{sec:baseline}, we implement the following three baseline models. The first one, denoted by~BL1, follows the idea of multi-stream from ICNet~\cite{zhao2018icnet}. The second one, denoted by~BL2, utilizes multi-level feature aggregation from FPN~\cite{lin2017feature} and U-Net~\cite{ronneberger2015u}. The last one, BL3, utilize the spatial pyramid pooling module proposed in PSPNet~\cite{zhao2017pyramid} with ResNet-101 as the backbone.

All the networks mentioned follow the same training strategy. We only use the 2925 fine annotated training images for training. The models are trained with the Adam optimizer~\cite{kingma2014adam} with initial learning rate $2\times 10^{-4}$, batch size $8$, momentum $0.9$, and weight decay $1\times 10^{-6}$. The `poly' learning rate policy is adopted with the power $0.9$. Data augmentation includes random flipping, mean subtraction, random scaling between [0.5, 2.0], and random cropping into 800 $\times$ 800 images.

The performances of baseline models are summarized in Table~\ref{table:baseline}. By default, we use BL2 as our baseline segmentation model in the following part.

\subsubsection{Using the Fast Feature Warping Module}
\label{ssub:ffw_eval}

To demonstrate the effectiveness of the FFW module, we compare the motion vector-based FFW module with the interpolation method and optical flow-based warping.
The interpolation method obtains the segmentation result of a certain frame by linearly interpolating the segmentation results of the previous and the next keyframe.
The optical flow-based warping, which takes optical flows instead of motion vectors as input, is similar to FFW, but it takes extra time for optical flow estimation.

\begin{figure}[tb]
\begin{center}
   \includegraphics[width=0.99\linewidth]{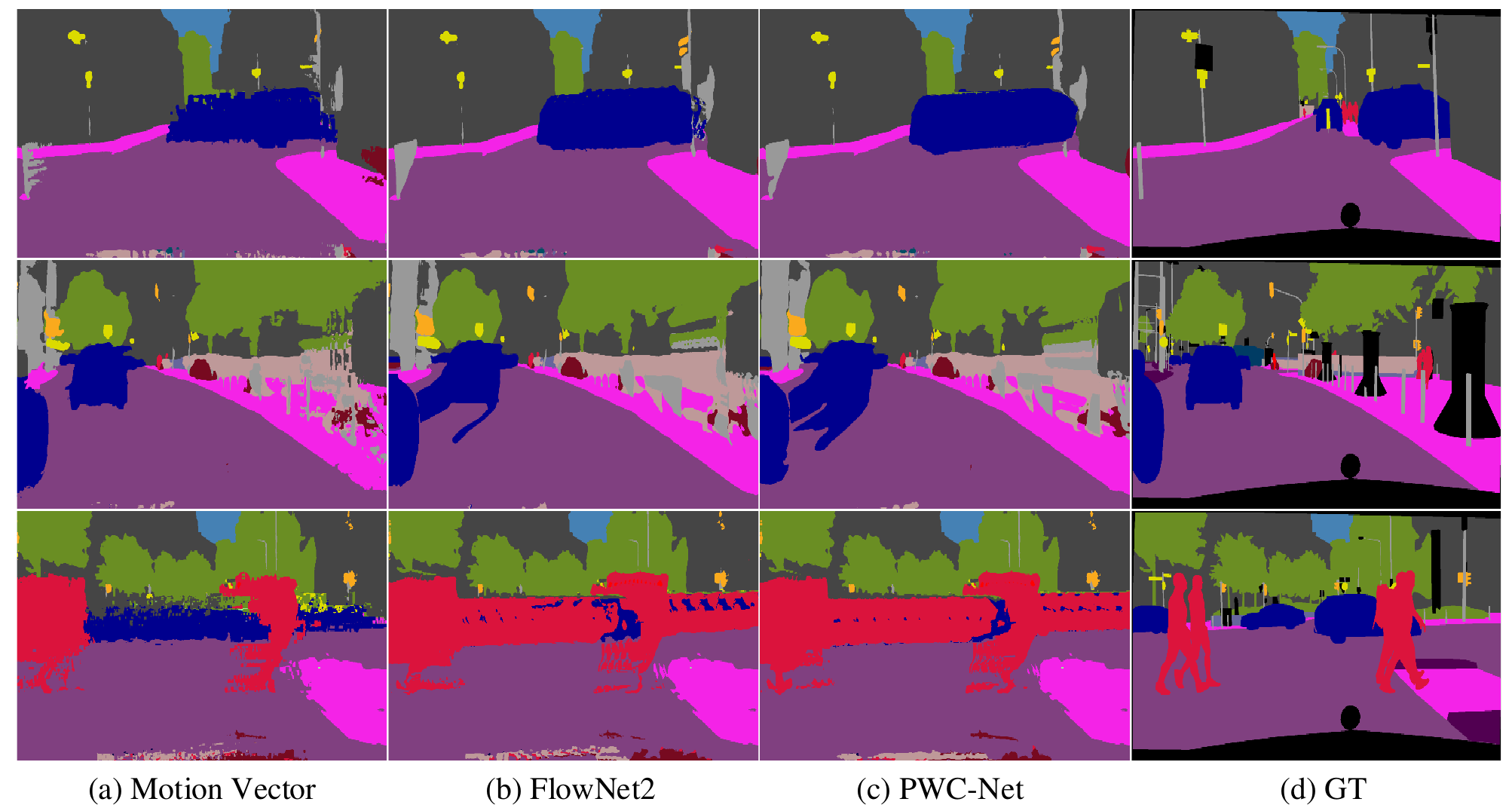}
\end{center}
	\vspace{-2em}
	\caption{Segmentation results w.r.t different kinds of motion inputs. The results using motion vectors are similar to those using optical flows.}
\label{fig:flowResult}
\end{figure}
\begin{figure}[b]
\begin{center}
   \includegraphics[width=1.0\linewidth]{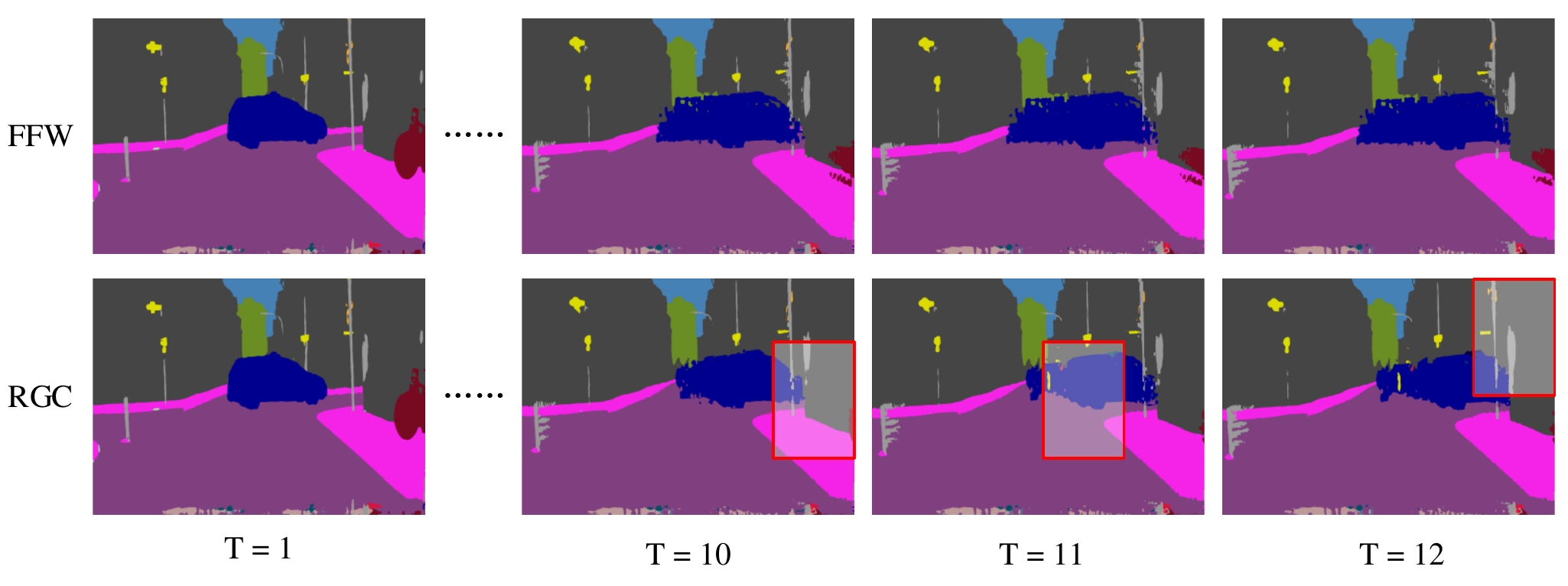}
\end{center}
	\vspace{-2em}
	\caption{Visualization of the RGC operation. The red rectangles are the regions selected by our RGC module. The results show that for the regions selected by the RGC module, the segmentation results are greatly improved. For example, for ``T=12'' in the figure, the boundaries of the pole are well-preserved by utilizing the RGC module.}
\label{fig:rgcvis}
\end{figure}
The comparison of these propagation methods is summarized in Table~\ref{table:mv}. \Cref{fig:flowResult} shows segmentation results w.r.t different kinds of flows. According to Table~\ref{table:mv} and Fig.~\ref{fig:flowResult}, the warped results of FlowNet2~\cite{ilg2017flownet} and PWC-Net~\cite{sun2018pwc} are not better than those of motion vectors. 
\ul{ We can see that the qualitative and quantitative results of adopting optical flows and motion vectors are similar. 
	We found that the segmentation accuracy is \mbox{\emph{not only}} attributed to the accuracy of the optical flow method we use. Interestingly, the key problem of warping-based segmentation methods is that they can only process the pixels which already exist in the previous frame, and therefore they can hardly deal with the \mbox{\emph{drastic or deformable movements}} of objects in the scene, which causes inaccurate predictions and makes the accuracy of every optical flow method drop to a relatively similar level. }

As shown in Table~\ref{table:mv}, both motion vector and optical flow-based warping achieves higher accuracy than interpolation. Compared with optical flow methods, FFW saves the time of flow estimation and achieves competitive accuracy. 
After applying FFW, all the three baseline models get several times of speed-up while the accuracy decreases to some degree.

In addition to increasing the speed, FFW unexpectedly performs better than baseline per-frame methods in some particular situations, as shown in Fig.~\ref{fig:ffwbt}. This is due to the moving of some objects through the boundaries of the camera view.  The per-frame method~(BL1) performs worse because it lacks the contextual information outside the camera view, whereas our FFW module can benefit from features extracted by previous frames.

	The results above are based on the configuration that the Group of Pictures (GOP) of a video is set to 12, the default value by MPEG-4. 
%
As shown in Equation~\ref{eq:time}, the average running time of TapLab is strongly correlated with the GOP number $g$. Fig.~\ref{fig:gop} illustrates accuracy~(mIoU) versus speed~(FPS) under different GOP configurations.

\subsubsection{Using the Region-Guided Correction Module} 

\begin{table}[b]
	\caption{Performance of Different Settings for the RGC Module \ul{on Cityscapes}} \label{table:rgc}
	\vspace{-1em}
  \begin{tabu} to 0.49\textwidth {X[l]cccc}
\toprule
Model&Stride&Region shape&mIoU&FPS\\
\midrule
BL1&-&-&67.3&33.2\\
BL1+FFW&-&-&60.6&169.5\\
BL1+FFW+RGC&256&(256, 256)&62.4&149.6\\
BL1+FFW+RGC&256&(256, 512)&62.6&146.8\\
BL1+FFW+RGC&512&(512, 512)&63.0&143.2\\
BL1+FFW+RGC&256&(512, 512)&63.4&137.5\\
\midrule
BL2+FFW+RGC&256&(512, 512)&68.2&131.4\\
\midrule
BL3+FFW+RGC&256&(512, 512)&72.2&33.8\\
\bottomrule
\end{tabu}

\medskip
\emph{\footnotesize `Region shape' indicates the height and width of sub-regions. `Stride' indicates the interval of sampled regions. These two parameters are only used for RGC.}
\end{table}

\begin{figure}[tb]
\begin{center}
   \includegraphics[width=1.0\linewidth]{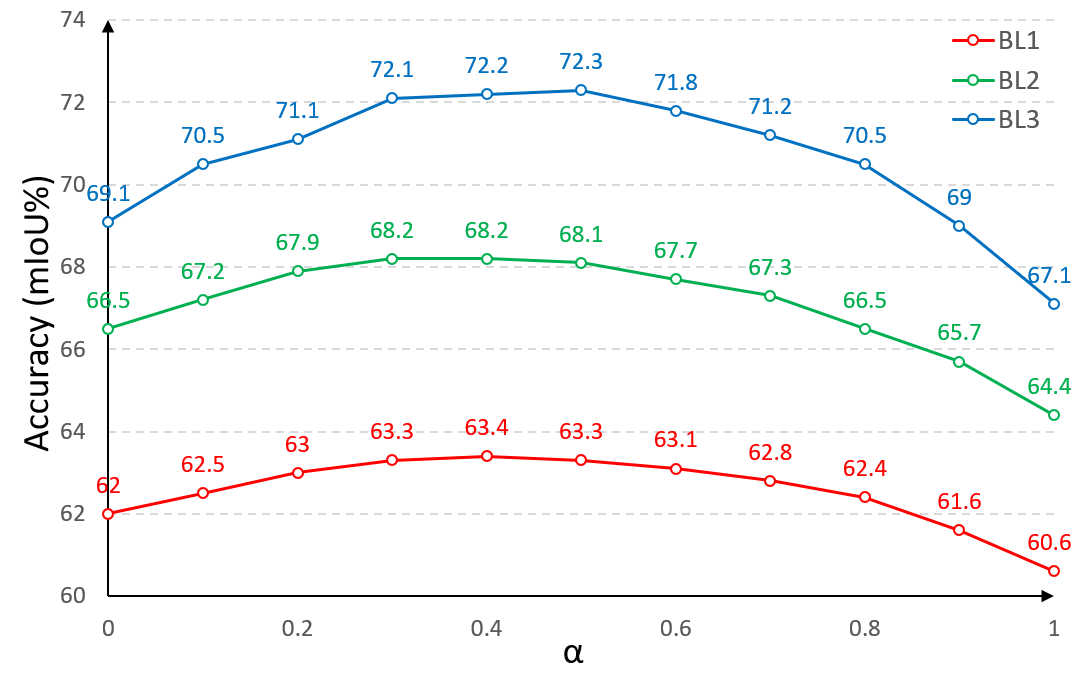}
\end{center}
	\vspace{-2em}
   \caption{Accuracy in different settings of $\alpha$. With the increase of $\alpha$, the accuracy first goes up and then goes down.}
\label{fig:alpha}
\end{figure}

Noticing the important role that residual maps play in video codec for motion compensation, we propose the residual-guided correction~(RGC) module to refine the propagated features. The correction procedure is shown in Fig.~\mbox{\ref{fig:rgcvis}}.
This improves the accuracy from $64.4\%$ to $68.2\%$~(BL2), as shown in Table~\ref{table:rgc}. 
Note that to alleviate the boundary-cropping problem, we set the ``stride'' parameter to keep the regions overlapped. When the stride is smaller than a region’s side(e.g., 256 v.s. 512), the candidate regions will be overlapping instead of adjacent so that even if a high-response object is sliced by the chosen region’s boundary, most of the object can stay in the region.
RGC can run in parallel with FFW to avoid extra running time. As shown in Table~\ref{table:rgc}, when the resolution of the input region is low enough, the inference speed grows disproportionately to the shrinking rate of the input shape, which means the dominator of inference time changes from computational costs to I/O and communication operations~(e.g., the time for `feed\_dict' in TensorFlow).

	Practically,  for the chosen region, we use the linear combination of warped feature maps,~$\mathcal{F}_w$, and the feature maps re-computed by the CNN,~$\mathcal{F}_{cnn}$, to form the final spatial feature maps, i.e.,
\begin{equation}
\begin{split}
	\mathcal{F} = (1-\alpha)\cdot \mathcal{F}_{w} + \alpha\cdot\mathcal{F}_{cnn},
\end{split}
\label{eq:lc}			
\end{equation}
	where $\alpha$ is the weight of combination. We study the effect of $\alpha$ as shown in Fig.~\ref{fig:alpha}. Notice that the feature maps directly obtained by the CNN, when $\alpha = 1$, do not achieve higher accuracy. We argue the concavity of this curve is caused by the following reasons. On the one hand, when $\alpha \rightarrow 1$ or $\mathcal{F}_{cnn}$ dominates, the small input region cannot capture enough global information, resulting in wrong predictions. On the other hand, when $\alpha \rightarrow 0$, the result feature maps are obtained from FFW with a lot of noise. Thus, only when $\alpha$ takes intermediate values, the result maps can take advantage of high responses from both. 

\begin{table}[tb]
	\caption{Effect of Each Module in TapLab \ul{on Cityscapes}}
\label{table:rgfs}
	\vspace{-1em}
  \begin{tabu} to 0.49\textwidth {X[1.5c]X[c]X[c]X[c]X[c]}
\toprule
FFW&RGC&RGFS&mIoU&FPS\\
\midrule
\textit{BL1}&&&&\\
$\checkmark$&&&60.6&\textbf{169.5}\\
$\checkmark$&$\checkmark$&&63.4&137.5\\
$\checkmark$&&$\checkmark$&63.8&123.5\\
$\checkmark$&$\checkmark$&$\checkmark$&\textbf{64.7}&106.9\\
\midrule
\textit{BL2}&&&&\\
$\checkmark$&&&64.4&\textbf{160.4}\\
$\checkmark$&$\checkmark$&&68.2&131.4\\
$\checkmark$&&$\checkmark$&69.4&114.0\\
$\checkmark$&$\checkmark$&$\checkmark$&\textbf{70.6}&99.8\\
\midrule
\textit{BL2 (PWC)}&&&&\\
$\checkmark$&&&61.7&\textbf{42.3}\\
$\checkmark$&$\checkmark$&&65.4&40.0\\
$\checkmark$&&$\checkmark$&67.4&40.7\\
$\checkmark$&$\checkmark$&$\checkmark$&\textbf{68.9}&38.7\\
\midrule
\textit{BL3}&&&&\\
$\checkmark$&&&67.1&\textbf{67.2}\\
$\checkmark$&$\checkmark$&&72.2&33.8\\
$\checkmark$&&$\checkmark$&72.0&38.3\\
$\checkmark$&$\checkmark$&$\checkmark$&\textbf{74.5}&25.4\\
\bottomrule
\end{tabu}

\medskip
\emph{\footnotesize FFW: Fast Feature Warping; RGC: Residual-Guided Correction; RGFS: Residual-Guided Frame Selection. ``$\checkmark$'' means the method utilizes this module.}
\end{table}

\begin{table}[t]
	\caption{Comparison of Different GOP Numbers}
	\vspace{-1.5em}
	\begin{center}
  \begin{tabularx}{0.42\textwidth}{cccc}
\toprule
GOP &Modules&mIoU&FPS\\
\midrule
\multirow{2}*{3}&FFW&71.9&72.2\\
&FFW+RGC+RGFS&72.5 (+0.6)&67.2\\
\midrule
\multirow{2}*{6}&FFW&69.3&114.0\\
&FFW+RGC+RGFS&71.5 (+2.2)&86.3\\
\midrule
\multirow{2}*{9}&FFW&67.4&141.2\\
&FFW+RGC+RGFS&71.0 (+3.6)&93.5\\
\midrule
\multirow{2}*{12}&FFW&64.4&160.4\\
&FFW+RGC+RGFS&70.6 (+6.2)&99.8\\
\bottomrule
\end{tabularx}
	\end{center}
	\vspace{-.5em}
\medskip
	\emph{\footnotesize The effect of RGC and RGFS in different settings of GOP numbers.}
\label{table:diffgop}
\end{table}

\begin{figure*}[t]
\begin{center}
   \includegraphics[width=0.9\linewidth]{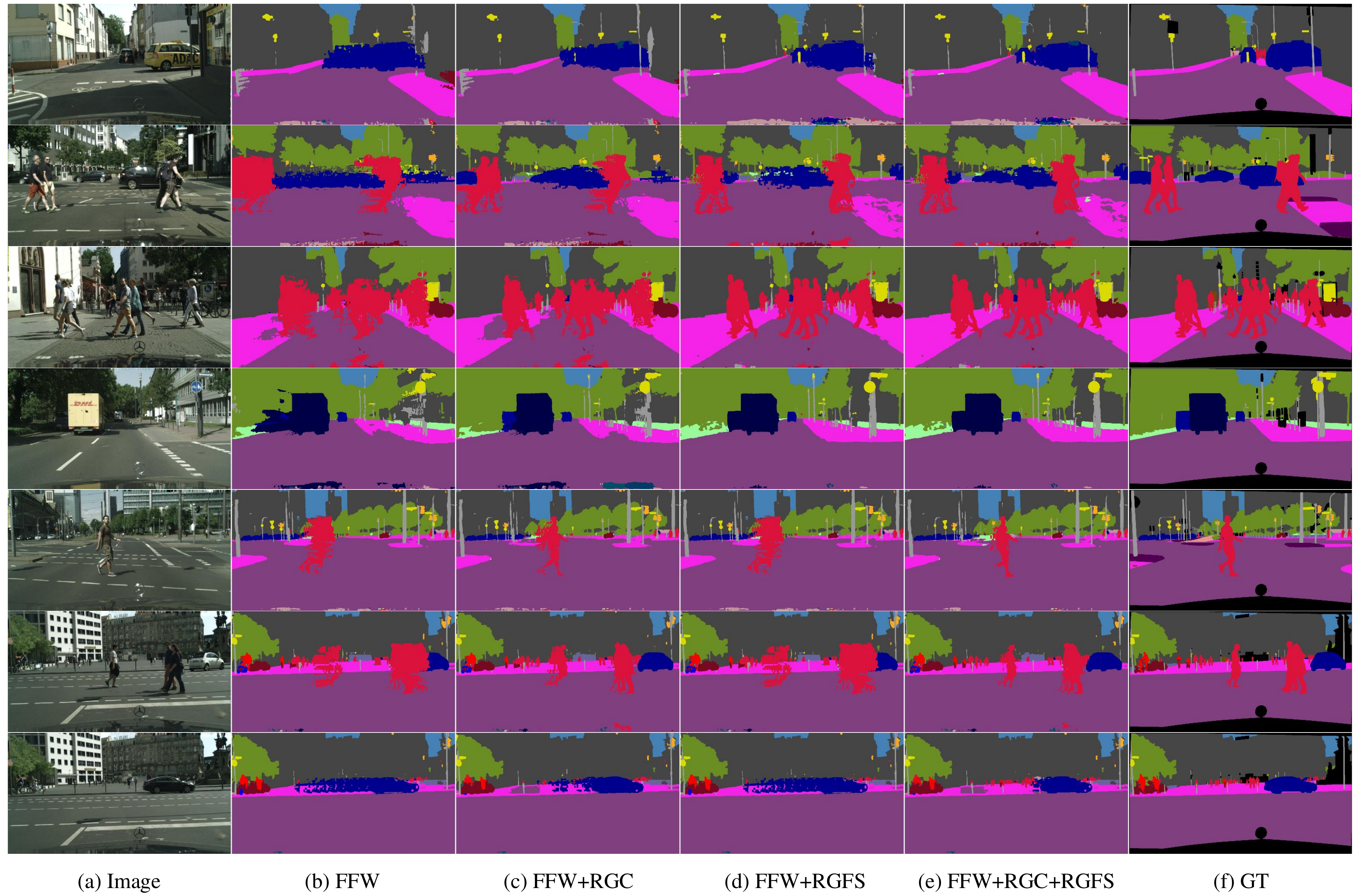}
\end{center}
	\vspace{-2em}
   \caption{Results on the Cityscapes val dataset. Despite the fact that FFW speeds up the segmentation, it also introduces noise in the results. With the help of RGC and RGFS, the boundaries of objects become much clearer.}
\label{fig:qlt}
\end{figure*}
\subsubsection{Using the Residual-Guided Frame Selection Module}

In addition to the correction of spatial regions, we also design the residual-guided frame selection~(RGFS) module to select the ``hard-to-warp'' P-frames and send them into the segmentation CNN. We set $\mathrm{THR_{RGFS}} = 3.6 \times 10^7 $ and this will approximately bring $10\%$ P-frames as keyframes.

As expected, this module further improves the segmentation accuracy from $68.2\%$ to $70.6\%$~(BL2). Table~\ref{table:rgfs} presents the effectiveness of different modules.  Notice that for BL1 and BL2, using RGC alone is faster than using RGFS alone while for BL3, it is the other way around. This is due to the slow BL3. It takes more time for BL3 to do region ($512\times 512$) segmentation for every single frame in RGC than to do full-size segmentation for 10\% P-frames in RGFS.

\ul{We also study the generality of RGC and RGFS by choosing PWC-Net instead of MV. As shown in Table~\mbox{\ref{table:rgfs}}, RGFS and RGC can consistently achieve better
performance for not only motion vectors but also optical flows. It is worth noting that residual maps corrects the corresponding imprecise motion vectors. They do not necessarily get along with optical flows. Thus, the accuracies of PWC-Net+RGC/RGFS are slightly lower than the corresponding MV-based versions.}

It is worth noting that our RGC and RGFS modules can be applied under all the GOP settings. As shown in Table~\mbox{\ref{table:diffgop}}, when the GOP number is large, the accuracy improves a lot while the speed may be much slower. When the GOP number is small, the accuracy gets improved with only a little more time consumed. To summarize, the RGC and RGFS modules are generic to different settings of GOP numbers.

\subsubsection{Qualitative Results}

The qualitative results of our framework on samples of Cityscapes are shown in Fig.~\ref{fig:qlt}. FFW speeds the process of segmentation but also introduces noise to the results. With the addition of RGC and RGFS, we obtain segmentation results with higher quality.

\subsection{Comparison with Other State-of-the-Art Methods}
\label{ssec:methodcmp}

\begin{table*}[tb]
  \caption{Comparison of Different Video Segmentation Methods on Cityscapes}
\label{table:cmp}
	\vspace{-1em}
\centering
  \begin{tabu} to 0.99\textwidth {X[2] X[c] X[c] X[c] X[2c] X[c] X[c]}
\toprule
Model&Eval set&mIoU&Resolution&GPU&FPS&FPS norm\\
\midrule
\textit{Per-frame Models}\\
  SegNet~\cite{badrinarayanan2017segnet}&val&57.0&256$\times$512&TITAN&16.7&36.3\\
  ENet~\cite{paszke2016enet}&test&58.3&512$\times$1024&TITAN X&76.9&79.3\\
  SQ~\cite{treml2016speeding}&test&59.8&1024$\times$2048&TITAN X(M)&16.7&27.4\\
  ICNet~\cite{zhao2018icnet}&val&67.7&1024$\times$2048&TITAN X(M)&30.3&49.7\\
  BiSeNet~\cite{yu2018bisenet}&val&69.0&768$\times$1536&TITAN Xp&105.8&98.8\\
  ESPNet~\cite{mehta2018espnet}&test&60.3&512$\times$1024&TITAN X&112&115.5\\
  ERFNet~\cite{romera2017erfnet}&test&68.0&512$\times$1024&TITAN X(M)&11.2&18.4\\
  DFANet~\cite{li2019dfanet}&test&71.3&1024$\times$1024&TITAN X&100&103.0\\
  SwiftNet~\cite{orsic2019defense}&val&70.4&1024$\times$2048&1080 Ti&39.9&39.9\\
\midrule
\midrule
\textit{Non-Per-Frame Models}\\
DFF~\cite{zhu2017deep}&val&69.2&512$\times$1024&Tesla K40&5.6&12.8\\
Low-Latency~\cite{li2018low}&val&75.89&1024$\times$2048&-&8.4&-\\
DVSNet1~\cite{xu2018dynamic}&val&63.2&1024$\times$2048&1080 Ti&30.4&30.4\\
DVSNet2~\cite{xu2018dynamic}&val&70.4&1024$\times$2048&1080 Ti&19.8&19.8\\
\midrule
\textit{TapLab}\\
BL2+FFW&val&64.4&1024$\times$2048&1080 Ti&160.4&160.4\\
BL2+FFW+RGC&val&68.2&1024$\times$2048&1080 Ti&131.4&131.4\\
BL2+FFW+RGFS&val&69.4&1024$\times$2048&1080 Ti&114.0&114.0\\
BL2+FFW+RGC+RGFS&val&70.6&1024$\times$2048&1080 Ti&99.8&99.8\\
BL2+FFW+RGC+RGFS&val&68.5&1024$\times$1024&1080 Ti&187.4&187.4\\
BL2+FFW+RGC+RGFS&val&69.9&768$\times$1536&1080 Ti&172.9&172.9\\
BL2+FFW+RGC+RGFS&test&69.2&1024$\times$2048&1080 Ti&99.8&99.8\\
\bottomrule
\end{tabu}

\medskip
  \emph{\footnotesize Results of semantic segmentation on Cityscapes. We select the best results of our models evaluated on the validation and compare them with previous works. We also report the inference speed, the input resolution, and the GPU platform. The default configurations of models are reported in Section~\ref{sssec:baseline} and Section~\ref{ssec:methodcmp}.}
\end{table*}

\begin{table}[tb]
	  \caption{Comparison of Different Models on CamVid}
		\vspace{-1em}
		\begin{center}
	  \begin{tabularx}{0.48\textwidth}{lccc}
	\toprule
	Model 						& mIoU	& FPS	& FPS-norm	\\
	\midrule
	ICNet [8]					& 67.1	& 27.8	& 45.6	\\
	BiSeNet	[9]					& 68.7	& -		& -	\\
	DFANet [10]					& 64.7	& 120	& 123.6	\\
	SwiftNet [11]				& 72.6	& -		& -	\\
	\midrule
	BL2							& 73.5	& 83.3	& 83.3	\\
	BL2+FFW						& 68.0	& 470.6	& 470.6 \\
	BL2+FFW+RGC					& 70.0	& 310.1	& 310.1 \\
	BL2+FFW+RGFS					& 70.4	& 327.9	& 327.9 \\
	BL2+FFW+RGC+RGFS				& 71.2	& 246.9	& 246.9 \\
	\bottomrule
	\end{tabularx}
		\end{center}
	\medskip
	\label{table:camvid}
\end{table}
	Finally, we compare our proposed framework with other state-of-the-art methods on Cityscapes validation set as shown in Table~\ref{table:cmp}.
We conduct all the experiments on a server with an Intel Core i7-6800K CPU and a single NVIDIA GeForce 1080 Ti GPU card. All our models run on the platform with CUDA 9.2, cuDNN 7.3 and TensorFlow 1.12. For a fair comparison, we follow the recent work of~\cite{orsic2019defense} and include the column ``FPS norm'', which provides a rough estimate on methods evaluated on other platforms and different resolutions. We use the scaling factors from the publicly available GPU benchmarks\footnote{\url{https://www.techpowerup.com/gpu-specs} and \url{https://github.com/jcjohnson/cnn-benchmarks}}. The scaling factors are 1.0 for GTX 1080 Ti, 1.07 for TITAN~Xp, 0.97 for TITAN~X~Pascal, 0.61 for TITAN~X~Maxwell, 0.46 for TITAN, and 0.44 for K40.

\ul{Note that TapLab is \textbf{not} bound to a specific baseline per-frame method. The baseline models used in our paper are representative but not carefully chosen. If a better per-frame model is adopted, the performance would be further improved.}

\subsection{Results on CamVid}
In this section, we provide qualitative and quantitative results on the CamVid dataset~\mbox{\cite{brostow2009semantic}}, which contains  video sequences at a resolution of 720\mbox{$\times$}960. We use the commonly used split, which partitions the dataset into 367 and 233 images for training and testing. During the evaluation, 11 semantic classes are taken into account.

The training protocol is the same as that of Cityscapes except for the crop size set to 600$\times$600, and we train the model for 20000 steps.
The threshold \mbox{$\mathrm{THR_{RGC}}$ is set to   $30$}. The threshold for frame selection \mbox{$\mathrm{THR_{RGFS}}$} is set to \mbox{$1.8 \times 10^7$} to keep 10\% P-frames selected by RGFS for full-resolution segmentation.

\begin{figure*}[tb]
\begin{center}
   \includegraphics[width=0.9\linewidth]{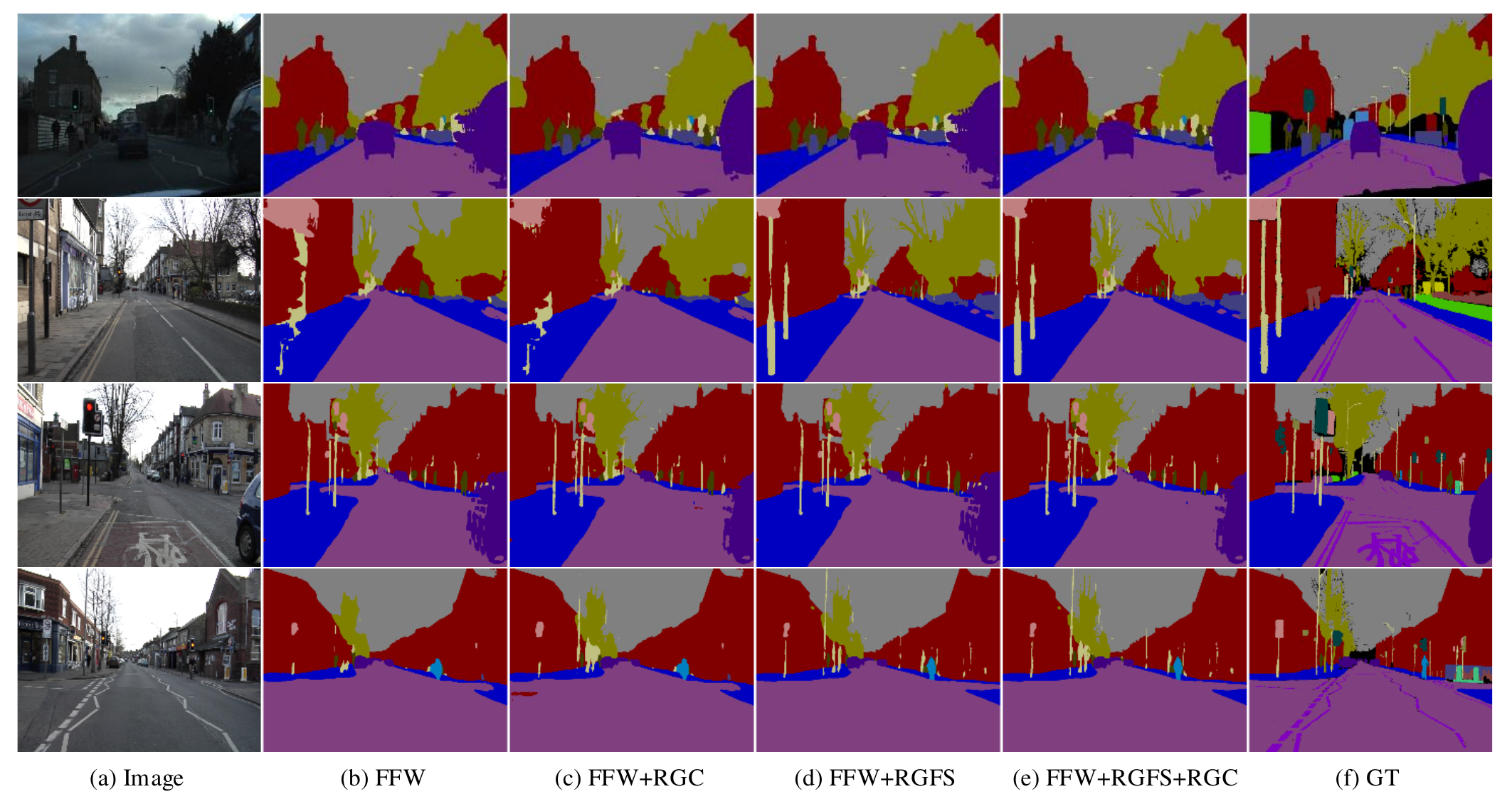}
\end{center}
	\vspace{-2em}
	\caption{Qualitative results on the CamVid dataset.  The results are consistent with those on Cityscapes.}
\label{fig:camvidVis}
\end{figure*}

Table~\mbox{\ref{table:camvid}} and Fig.~\mbox{\ref{fig:camvidVis}} show the quantitative and qualitative results of TapLab on CamVid. Without loss of generality, we use BL2 as the baseline model. According to the results, our TapLab achieves consistent results on this dataset.
Note that the changes between adjacent frames are slight, since the frequency of videos in CamVid~(30 Hz) is higher than that in Cityscapes~(17 Hz). Thus, the accuracy degradation incurred by applying warping is smaller.

\section{Conclusion}
\label{sec:conclusion}

In this paper, we present a novel compressed feature-based framework to perform semantic video segmentation effectively. It incorporates a fast feature warping module, a residual-guided correction module, and a residual-guided frame selection module as key components to strike a balance between accuracy and speed. The modules are generic to most kinds of existing CNNs for segmentation, and they can easily be added or not to meet the actual hardware requirements. \ul{The experimental results on Cityscapes and CamVid demonstrate that our framework significantly speed up various types of per-frame segmentation models.} In the future, we will explore more ways to utilize compressed-domain features to improve accuracy.

\ifCLASSOPTIONcompsoc
  \section*{Acknowledgments}
\else
  \section*{Acknowledgment}
\fi
Junyi Feng and Songyuan Li contributed equally to this work.


\ifCLASSOPTIONcaptionsoff
  \newpage
\fi



%
\bibliographystyle{IEEEtran}
\bibliography{bibli}
%
%

%

%
\begin{IEEEbiography}[{\includegraphics[width=1in,height=1.25in,clip,keepaspectratio]{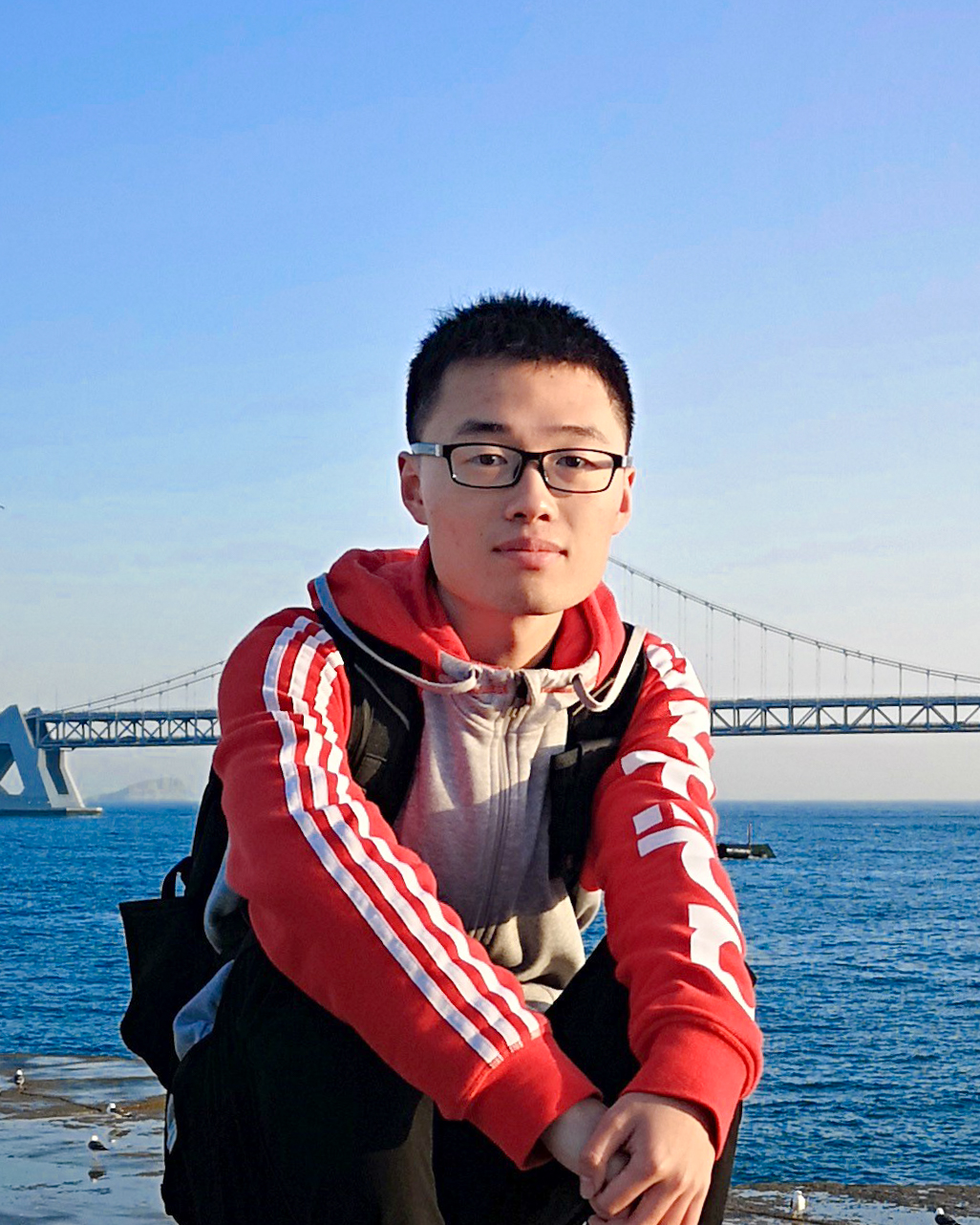}}]{Junyi Feng}
received his B.E. degree in 2018 from ShanghaiTech University, Shanghai, China, where he worked on problems in computer vision and computer graphics. He is currently a Master Student at College of Computer Science and Technology, Zhejiang University. His current research interests include semantic segmentation, neural network for video processing, and object detection.
\end{IEEEbiography}
\vspace{-8em}

\begin{IEEEbiography}[{\includegraphics[width=1in,height=1.25in,clip,keepaspectratio]{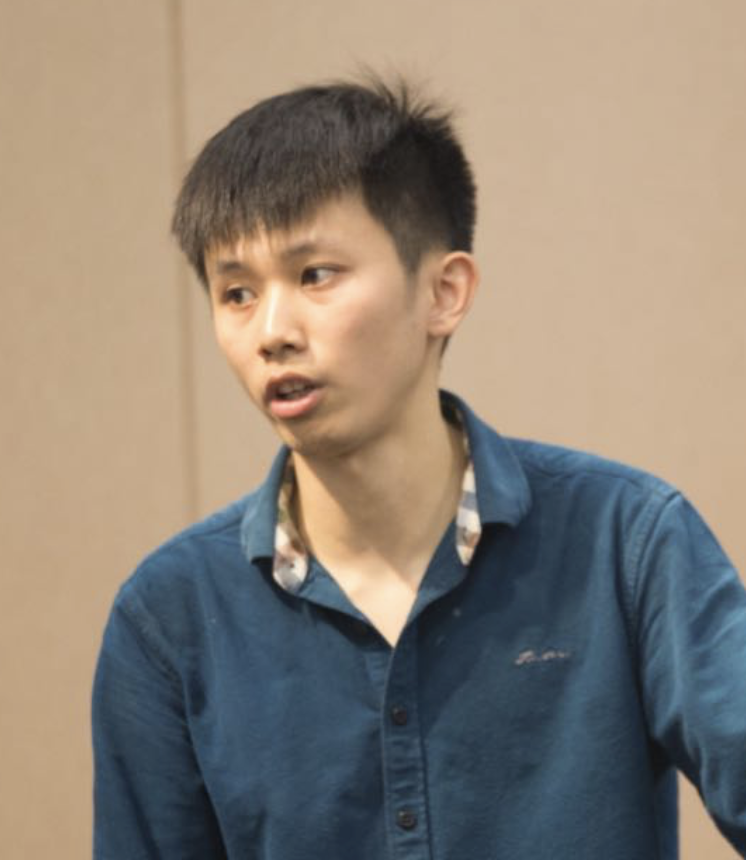}}]{Songyuan Li}
received his master's degree in 2017 from Zhejiang University, China, where he worked on problems in computer architecture and operating systems. He is currently a Ph.D. candidate at Zhejiang University. His current research interests include semantic segmentation and video localization.
\end{IEEEbiography}
\vspace{-8em}
%
%
\begin{IEEEbiography}[{\includegraphics[width=1in,height=1.25in,clip,keepaspectratio]{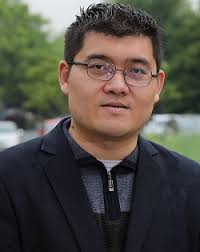}}]{Xi Li}
received the Ph.D. degree from the National Laboratory of Pattern
Recognition, Chinese Academy of Sciences, Beijing, China, in 2009. From
2009 to 2010, he was a Post-Doctoral Researcher with CNRS Telecom ParisTech, France. He was a Senior Researcher with the University of
Adelaide, Australia. He is currently a Full Professor with Zhejiang University,
China. His research interests include visual tracking, motion analysis, face
recognition, Web data mining, and image and video retrieval.
\end{IEEEbiography}
\vspace{-8em}

\begin{IEEEbiography}[{\includegraphics[width=1in,height=1.25in,clip,keepaspectratio]{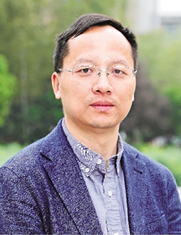}}]{Fei Wu}
received the B.S. degree from Lanzhou
University, Lanzhou, Gansu, China, the M.S. degree
from Macao University, Taipa, Macau, and the Ph.D.
degree from Zhejiang University, Hangzhou, China.
He is currently a Full Professor with the College
of Computer Science and Technology, Zhejiang
University. He was a Visiting Scholar with Prof.
B. Yu’s Group, University of California, Berkeley,
from 2009 to 2010. His current research interests
include multimedia retrieval, sparse representation,
and machine learning.
\end{IEEEbiography}

\begin{IEEEbiography}[{\includegraphics[width=1in,height=1.25in,clip,keepaspectratio]{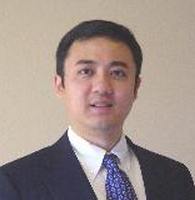}}]{Qi Tian}
received the B.E. degree in electronic engineering from Tsinghua University, the M.S. degree in ECE from Drexel University, and the Ph.D. degree in ECE from the University of Illinois at Urbana–Champaign (UIUC). He is currently the Chief Scientist in computer vision with Huawei Noah’s Ark Lab. He is also on a faculty leave and a Full Professor with the Department of Computer Science, The University of Texas at San Antonio (UTSA). From 2008 to 2009, he took one-year faculty leave at Microsoft Research Asia (MSRA). He has published over 480 refereed journal and conference articles. His Google Citation is over 15 200 with H-index 64. He was the coauthor of best articles, including ICME 2019, CIKM 2018, ACM ICMR 2015, PCM 2013, MMM 2013, and ACM ICIMCS 2012, a Top 10\% Paper Award in MMSP 2011, a Student Contest Paper in ICASSP 2006, and the coauthor of a Best Paper/Student Paper Candidate in ICME 2015 and PCM 2007. His research projects are funded by ARO, NSF, DHS, Google, FXPAL, NEC, SALSI, CIAS, Akiira Media Systems, HP, Blippar, and UTSA. His research interests include computer vision, multimedia information retrieval, and machine learning. He received the 2017 UTSA Presidents Distinguished Award for Research Achievement, the 2016 UTSA Innovation Award, the 2014 Research Achievement Awards from College of Science, UTSA, the 2010 Google Faculty Award, and the 2010 ACM Service Award. He is an Associate Editor of the IEEE TMM, TNNLS, TCSVT, ACM TOMM, and MMSJ and on the Editorial Board of the Journal of Multimedia (JMM) and the Journal of Machine Vision Applications (MVA). He is also the Guest Editor of the IEEE TMM and Journal of CVIU.
\end{IEEEbiography}

\begin{IEEEbiography}[{\includegraphics[width=1in,height=1.25in,clip,keepaspectratio]{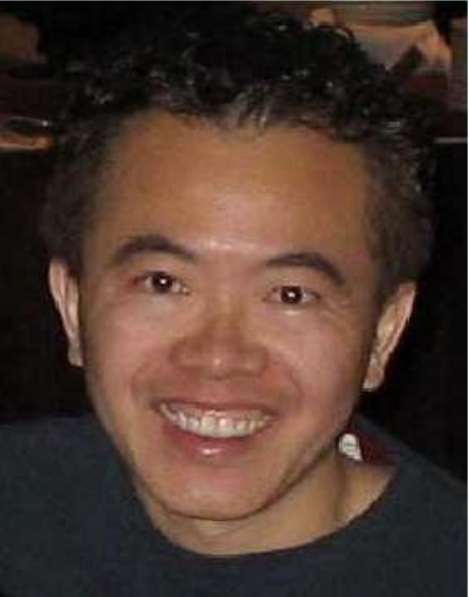}}]{Ming-Hsuan Yang}
received the PhD degree in computer science from the University of Illinois at Urbana-Champaign, in 2000. He is a professor in electrical engineering and computer science with the University of California, Merced. He served as an associate editor of the IEEE Transactions on Pattern Analysis and Machine Intelligence from 2007 to 2011, and is an associate editor of the International Journal of Computer Vision, Image and Vision Computing and the Journal of
Artificial Intelligence Research. He received the NSF CAREER award in 2012 and the Google Faculty Award in 2009. He is a senior member of the IEEE and ACM.
\end{IEEEbiography}

\begin{IEEEbiography}[{\includegraphics[width=1in,height=1.25in,clip,keepaspectratio]{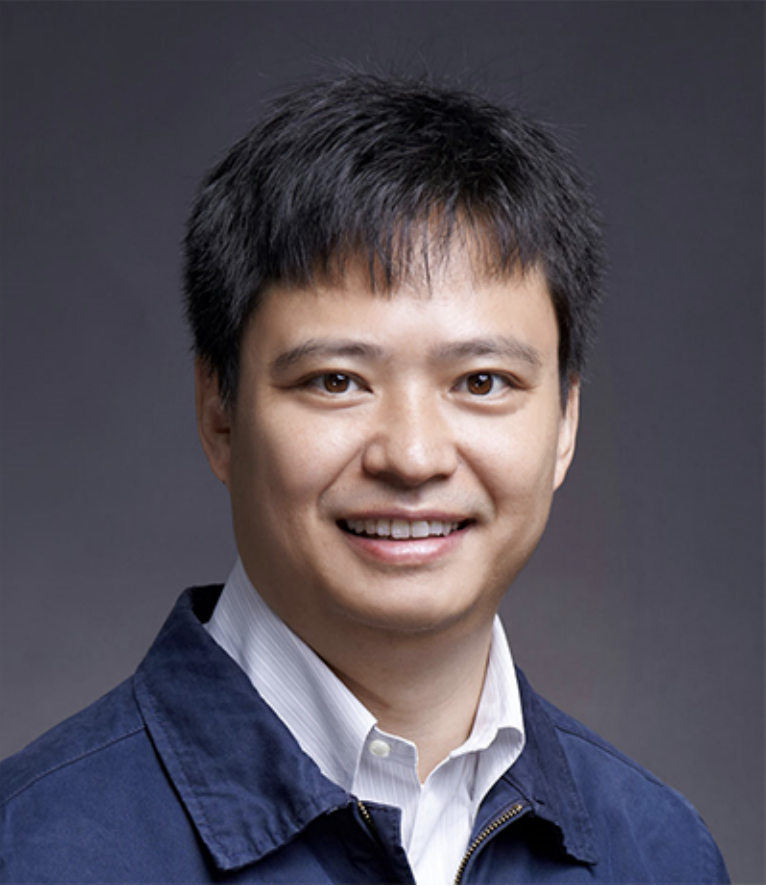}}]{Haibin Ling}
received the B.S. and M.S. degrees from Peking University in 1997 and 2000, respectively, and the Ph.D. degree from the University of Maryland, College Park, in 2006. From 2000 to 2001, he was an assistant researcher at Microsoft Research Asia. From 2006 to 2007, he worked as a postdoctoral scientist at the University of California Los Angeles. In fall 2019, he will join the Computer Science Department of Stony Brook University as SUNY Empire Innovation Professor. He received the Best Student Paper Award at ACM UIST in 2003, NSF CAREER Award in 2014, and Amazon Research Award in 2019. He is an Associate Editor of IEEE Trans. on Pattern Analysis and Machine Intelligence, Pattern Recognition, and Computer Vision and Image Understanding, and has served as Area Chairs for CVPR several times.
\end{IEEEbiography}




\end{document}